\documentclass[twocolumn]{svjour3} 
\smartqed  

\usepackage{graphicx}
\usepackage[numbers]{natbib}
\usepackage{multirow}
\usepackage{amsmath}
\usepackage{amssymb}
\usepackage{bbm}
\usepackage{bm}
\usepackage{booktabs}
\usepackage{array} 
\usepackage{blindtext}
\usepackage{comment}

\usepackage{pifont}

\usepackage[dvipsnames]{xcolor}
\usepackage{color, colortbl}
\definecolor{citecolor}{HTML}{0071bc}
\definecolor{tabhighlight}{HTML}{e5e5e5}
\usepackage[colorlinks,citecolor=citecolor]{hyperref}

\makeatletter
\renewcommand\paragraph{
  \@startsection{paragraph} 
  {4} 
  {\z@} 
  {.5em \@plus1ex \@minus.2ex} 
  {-.5em} 
  {\normalfont\normalsize\bfseries} 
}
\makeatother

\def\eg{\emph{e.g.}}
\def\ie{\emph{i.e.}}
\def\etal{\emph{et al.}}

\def\wrt{w.r.t.}
\def\aka{\emph{a.k.a.}}
\def\wrt{\emph{w.r.t. }}
\def\vs{\emph{v.s. }}
\newcommand{\method}[1]{\noindent \textbf{#1.}}
\usepackage[misc]{ifsym}

\usepackage{amsmath,amssymb,dsfont}

\usepackage{titlesec}
\titlespacing*{\section}
{0pt}{1.1ex plus 1ex minus .2ex}{1.1ex plus .2ex}
\titlespacing*{\subsection}
{0pt}{0.6ex plus 1ex minus .2ex}{0.6ex plus .2ex}

\definecolor{revision}{HTML}{DC143C}
\newcommand{\tim}[1]{{\color{revision}#1}}

\begin{document}
\sloppy

\title{A Comprehensive Survey on Test-Time Adaptation under Distribution Shifts}
\author{Jian Liang $^{1,2}$ \and Ran He $^{1,2}$ \and Tieniu Tan $^{1,2,3}$}

\institute{
$^1$ Center for Research on Intelligent Perception and Computing \& State Key Laboratory of Multimodal Artificial Intelligence Systems, CASIA, China \\
$^2$ School of Artificial Intelligence, University of Chinese Academy of Sciences, China \\
$^3$ Nanjing University, China\\
\tt{\Letter\ \href{mailto:liangjian92@gmail.com}{liangjian92@gmail.com}, \{rhe,tnt\}@nlpr.ia.ac.cn}
}

\date{Received: date / Accepted: date}

\maketitle

\begin{abstract}
Machine learning methods strive to acquire a robust model during the training process that can effectively generalize to test samples, even in the presence of distribution shifts.
However, these methods often suffer from performance degradation due to unknown test distributions.
Test-time adaptation (TTA), an emerging paradigm, has the potential to adapt a pre-trained model to unlabeled data during testing, before making predictions.
Recent progress in this paradigm has highlighted the significant benefits of using unlabeled data to train self-adapted models prior to inference.
In this survey, we categorize TTA into several distinct groups based on the form of test data, namely, test-time domain adaptation, test-time batch adaptation, and online test-time adaptation.
For each category, we provide a comprehensive taxonomy of advanced algorithms and discuss various learning scenarios.
Furthermore, we analyze relevant applications of TTA and discuss open challenges and promising areas for future research.
For a comprehensive list of TTA methods, kindly refer to \url{https://github.com/tim-learn/awesome-test-time-adaptation}.
\end{abstract}

\section{Introduction} \label{sec:intro}
Traditional machine learning methods assume that the training and test data are drawn independently and identically (i.i.d.) from the same distribution \citep{quinonero2008dataset}.
However, when the test distribution (target) differs from the training distribution (source), we face the problem of \emph{distribution shifts}.
Such a shift poses significant challenges for machine learning systems deployed in the wild, such as images captured by different cameras \citep{saenko2010adapting}, road scenes of different cities \citep{chen2017no}, and imaging devices in different hospitals \citep{liu2022source}.
As a result, the research community has developed a variety of generalization or adaptation techniques to improve model robustness against distribution shifts. 
For instance, \emph{domain generalization} (DG) \citep{zhou2022dg} aims to learn a model using data from one or multiple source domains that can generalize well to any out-of-distribution target domain. 
On the other hand, \emph{domain adaptation} (DA) \citep{kouw2019review} follows the transductive learning principle to leverage knowledge from a labeled source domain to an unlabeled target domain.

This survey primarily focuses on the paradigm of \emph{test-time adaptation} (TTA), which involves adapting a pre-trained model from the source domain to unlabeled data in the target domain before making predictions \citep{liang2020we,sun2020test,wang2021tent}. 
While DG operates solely during the training phase, TTA has the advantage of being able to access test data from the target domain during the test phase. 
This enables TTA to enhance recognition performance through adaptation with the available test data.
Additionally, DA typically necessitates access to both labeled data from the source domain and (unlabeled) data from the target domain simultaneously, which can be prohibitive in privacy-sensitive applications such as medical data.
In contrast, TTA only requires access to the pre-trained model from the source domain, making it a secure and practical alternative solution.

\begin{figure*}[tbp]
\centering
\begin{tabular}{ccc} 
 \includegraphics[trim={0.0cm 0.0cm 0.0cm 0.0cm},clip, width=0.32\textwidth]{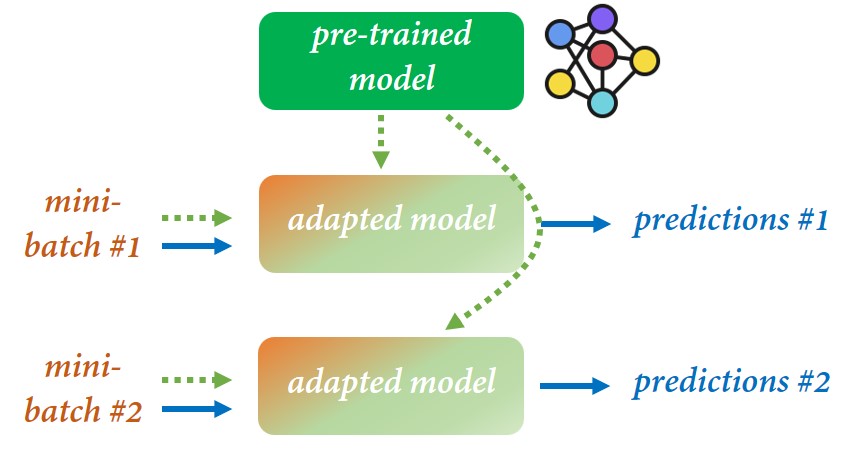} & \includegraphics[trim={0.0cm 0.0cm 0.0cm 0.0cm},clip, width=0.32\textwidth]{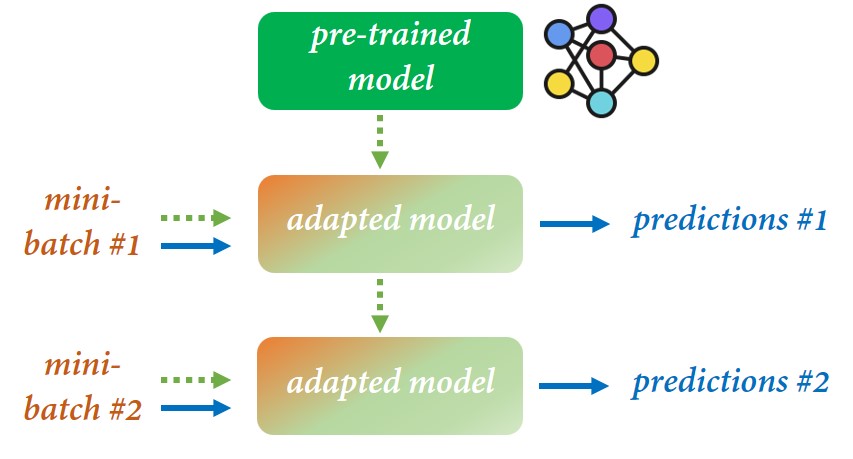} & \includegraphics[trim={0.0cm 0.0cm 0.0cm 0.0cm},clip, width=0.32\textwidth]{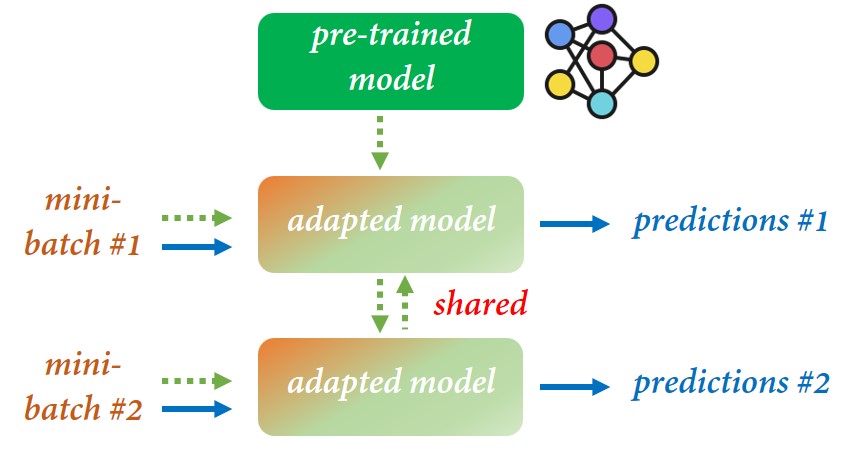} \\ 
 (a) test-time batch adaptation & (b) online test-time adaptation & (c) test-time domain adaptation \\ 
\end{tabular}
\caption{The \textbf{test-time adaptation (TTA) paradigm} aims to adapt the pre-trained model to various types of unlabeled test data, including \tim{single mini-batch in (a), streaming data in (b), or an entire dataset in (c)}, before making predictions. During the adaptation process, either the model or the input data can be altered to improve performance against distribution shifts. \tim{The dotted green arrow indicates the test-time training phase before inference, while the blue arrow denotes pure inference.}}
\label{fig:task}
\end{figure*}

Based on the characteristics of the test data \footnote{In this survey, we use the terms ``test data" and ``target data" interchangeably to refer to the data used for adaptation at test time.}, TTA methods can be categorized into three distinct cases in Fig.~\ref{fig:task}: \emph{test-time domain adaptation} (TTDA), \emph{test-time batch adaptation} (TTBA), and \emph{online test-time adaptation} (OTTA).
For a better illustration, let us consider a scenario where there are $m$ unlabeled mini-batches denoted as ${b_1, \cdots, b_m}$ during test time.
Firstly, TTDA, also known as source-free domain adaptation \citep{liang2020we,kundu2020universal,li2020model}, utilizes all $m$ test batches for multi-epoch adaptation before generating final predictions. 
Secondly, TTBA individually adapts the pre-trained model to one \footnote{Such a single-sample adaptation corresponds to a batch size of 1, \aka, test-time instance adaptation.} or a few instances \citep{sun2020test,zhang2022memo,schneider2020improving,zhang2021adaptive}.
In other words, the predictions made for each mini-batch are independent of the predictions made for the other mini-batches.
Thirdly, OTTA \citep{wang2021tent,iwasawa2021test,wang2022continual} adapts the pre-trained model to the target data $\{b_1, \cdots, b_m\}$ in an online manner, where each mini-batch can only be observed once.
Importantly, the knowledge learned from previously observed mini-batches can facilitate adaptation to the current mini-batch.
It is worth emphasizing that OTTA methods can be applied to TTDA with multiple epochs, and TTBA methods can be applied to OTTA with the assumption of knowledge reuse.

In this survey, we for the first time define the broad concept of \emph{test-time adaptation} and consider the three aforementioned topics (\ie, TTDA, TTBA, and OTTA) as its special cases.
Subsequently, we thoroughly review the advanced algorithms for each topic and present a summary of various applications related to TTA.
Our contributions can be summarized into three key aspects.
\begin{enumerate}
\item To our knowledge, this is the first survey that provides a systematic overview of three distinct topics within the broad test-time adaptation paradigm.
\item We propose a novel taxonomy of existing methods and provide a clear definition for each topic. 
We hope this survey will help readers gain a deeper understanding of the advancements in each area.
\item We analyze various applications related to the TTA paradigm in Sec.~\ref{sec:appl}, and provide an outlook of recent emerging trends and open problems in Sec.~\ref{sec:future} to shed light on future research directions.
\end{enumerate}

\method{Comparison with previous surveys} 
While our survey contributes to the broader research area of DA, which has been previously reviewed in other works such as \citep{kouw2019review,wilson2020survey}, our specific focus is on test-time adaptation where the availability of source data during adaptation is limited or non-existent.
Two recent surveys \citep{fang2024source,li2024comprehensive} have focused on source-free domain adaptation which is a particular topic extremely similar to TTDA discussed in our survey.
Even within the specific topic, we provide a novel taxonomy that encompasses a wider range of related papers.
Another survey \citep{liu2021data} considers source-free domain adaptation as an instance of data-free knowledge transfer, which shares some overlap with our survey.
However, we unify TTDA and several related topics from the perspective of model adaptation under distribution shifts.
We believe that it is a novel and pivotal contribution to the field of transfer learning.

\section{Related Research Topics}
\label{sec:related}
\subsection{Domain Adaptation}
As a special case of transfer learning \citep{pan2009survey}, DA \citep{ben2010theory} typically leverages labeled data from a source domain to learn a classifier for an unlabeled target domain with a different distribution, in a transductive learning manner \citep{joachims1999transductive}.
There are two major assumptions of distribution shift \citep{quinonero2008dataset}: \emph{covariate shift} in which the input features cause the labels; and \emph{label shift} in which the output labels cause the features.
We briefly introduce a few popular techniques and refer the reader to the existing literature on DA (\eg, \citep{kouw2019review,wilson2020survey}) for further information.
DA methods rely on the existence of source data to bridge the domain gap, and existing techniques can be broadly divided into four categories, \ie, input-level translation \citep{bousmalis2017unsupervised,hoffman2018cycada}, feature-level alignment \citep{long2015learning,ganin2015unsupervised,tzeng2017adversarial}), output-level regularization \citep{chen2019domain,cui2020towards,jin2020minimum}, and class-prior estimation \citep{saerens2002adjusting,lipton2018detecting,azizzadenesheli2019regularized}.
If it is feasible to generate training data from the source model \citep{li2020model}, then the task of TTDA can be tackled using conventional DA methods.
Likewise, one relevant topic closely related to TTBA (batch size equals 1) is \emph{one-shot domain adaptation} \citep{luo2020adversarial,varsavsky2020test}, which entails adapting to a single unlabeled instance while still necessitating the source domain during adaptation.
Moreover, OTTA is closely related to \emph{online domain adaptation} \citep{moon2020multi,yang2022burn}, which involves adapting to an unlabeled target domain with streaming data that is promptly deleted after adaptation.

\subsection{Hypothesis Transfer Learning}
\emph{Hypothesis transfer learning} (HTL) \citep{kuzborskij2013stability} is another special case of transfer learning where pre-trained models (source hypotheses) retain information about previously encountered tasks.
Shallow HTL methods \citep{yang2007cross,tommasi2013learning,ahmed2020camera} typically assume that the optimal target hypothesis is closely associated with these source hypotheses, and subsequent methods \citep{ao2017fast,nelakurthi2018source} extend this approach to a semi-supervised scenario where unlabeled target data are also utilized for training.
Fine-tuning \citep{yosinski2014transferable} is a typical example of a deep HTL method that may update a partial set of parameters in the source model.
Despite HTL methods assuming no explicit access to the source domain or any knowledge about the relatedness of the source and target distributions, they still require a certain quantity of labeled data in the target domain.
Another related topic is \emph{domain-incremental learning} \citep{van2022three,wang2022s} which tackles the same type of problem but in diverse contexts.
However, such an incremental learning task focuses more on the anti-forgetting ability after learning a supervised task.

\subsection{Domain Generalization}
DG \citep{li2018learning,carlucci2019domain,gulrajani2020search} aims to learn a model from one or multiple different but related domains that can generalize well on unseen testing domains. 
Researchers often devise specialized training techniques to enhance the generalization capability of the pre-trained model, which can be compatible with the studied TTA paradigm.
Notably, MAML \citep{finn2017model} is a representative approach that learns the initialization of a model's parameters to achieve optimal fast learning on a new task using a small number of samples and gradient steps.
Such a meta-learning strategy offers a straightforward solution for TTA without the incorporation of test data in the meta-training stage.
For further information, we refer the reader to existing literature (\eg, \citep{zhou2022dg,wang2022generalizing,hospedales2021meta}).

\subsection{Self-Supervised Learning} 
\emph{Self-supervised learning} \citep{jing2020self} is a learning paradigm that focuses on how to learn from unlabeled data by obtaining supervisory signals from the data itself through pretext tasks that leverage its underlying structure. 
Early pretext tasks in the computer vision field include image colorization \citep{zhang2016colorful}, image inpainting \citep{pathak2016context}, and image rotation \citep{gidaris2018unsupervised}. 
Advanced pretext tasks like clustering \citep{caron2018deep,caron2020unsupervised} and contrastive learning \citep{he2020momentum,chen2020simple} have achieved remarkable success, even exceeding the performance of their supervised counterparts.
Self-supervised learning is also popular in other fields like natural language processing \citep{kenton2019bert}, speech processing \citep{baevski2020wav2vec}, and graph-structured data \citep{you2020graph}.
For TTA tasks, these self-supervised learning techniques can be utilized to help learn discriminative features \citep{liang2021source} or act as an auxiliary task \citep{sun2020test}. 

\subsection{Semi-Supervised Learning} 
\emph{Semi-supervised learning} \citep{chen2022semi} is another learning paradigm concerned with leveraging unlabeled data to reduce the reliance on labeled data.
A common objective for semi-supervised learning methods comprises two terms: a supervised loss over labeled data and an unsupervised loss over unlabeled data.
Regarding the latter term, there are three typical cases: self-training \citep{grandvalet2004semi,lee2013pseudo}, which encourages the model to produce confident predictions; consistency regularization under input variations \citep{miyato2018virtual,sohn2020fixmatch} and model variations \citep{laine2017temporal,tarvainen2017mean}, which forces networks to output similar predictions when inputs or models are perturbed; and graph-based regularization \citep{iscen2019label}, which seeks local smoothness by maximizing the pairwise similarities between nearby data points.
For TTA tasks, these semi-supervised learning techniques can be easily integrated to unsupervisedly update the pre-trained model during adaptation.

\subsection{Test-Time Augmentation}
\emph{Test-time augmentation} \citep{shanmugam2021better} employs data augmentation techniques \citep{shorten2019survey} (\eg, geometric transformations and color space augmentations) on test images to boost prediction accuracy \citep{he2016deep}, estimate uncertainty \citep{smith2018understanding}, and enhance robustness \citep{guo2018countering,perez2021enhancing}.
As a typical example, ten-crop testing \citep{he2016deep} computes the final prediction by averaging predictions from ten different scaled versions of a test image.
Other popular aggregation strategies include selective augmentation \citep{kim2020learning} and learnable aggregation weights \citep{shanmugam2021better}.
In addition to data variation, Monte Carlo (MC) dropout \citep{gal2016dropout} enables dropout within the network during testing and performs multiple forward passes with the same input data to estimate the model uncertainty.
Generally, test-time augmentation techniques do not explicitly consider distribution shifts but can be advantageous for TTA methods.

\section{Test-Time Domain Adaptation}
\label{sec:ttda}
\subsection{Problem Definition}
\begin{definition}[Domain]
A domain $\mathcal{D}$ is a joint distribution $p(x,y)$ defined on the input-output space $\mathcal{X} \times \mathcal{Y}$, where random variables $x \in \mathcal{X}$ and $y \in \mathcal{Y}$ denote the input data and the label (output), respectively.
\end{definition}

In a well-studied DA problem, the domain of interest is called the target domain $p_\mathcal{T}(x,y)$ and the domain with labeled data is called the source domain $p_\mathcal{S}(x,y)$.
The label $y$ can either be discrete (in a classification task) or continuous (in a regression task).
Unless otherwise specified, $\mathcal{Y}$ is a $C$-cardinality label set, and we usually have one labeled source domain $\mathcal{D}_\mathcal{S}=\{(x_1,y_1),\dots,(x_{n_s},y_{n_s})\}$ and one unlabeled target domain $\mathcal{D}_\mathcal{T}=\{x_1,\dots,x_{n_t}\}$ under data distribution shifts: $\mathcal{X}_\mathcal{S}=\mathcal{X}_\mathcal{T}, p_\mathcal{S}(x) \not= p_\mathcal{T}(x)$, including the \emph{covariate shift} \citep{quinonero2008dataset} assumption ($p_\mathcal{S}(y|x) = p_\mathcal{T}(y|x)$).
Other distribution shifts like \emph{prior shift} \citep{saerens2002adjusting} are further discussed in Sec. \ref{sec:ttda-task}.
Typically, the unsupervised domain adaptation (UDA) paradigm aims to leverage supervised knowledge in $\mathcal{D}_\mathcal{S}$ to help infer the label of each target sample in $\mathcal{D}_\mathcal{T}$.

Chidlovskii \etal \citep{chidlovskii2016domain} for the first time consider performing domain adaptation with no access to source data.
Specifically, they propose three scenarios for feature-based domain adaptation with: source classifier with accessible models and parameters, source classifier as a black-box model, and source class means as representatives.
This new setting utilizes all the test data to adjust the classifier learned from the training data, which could be regarded as a broad test-time adaptation scheme. 
Several methods \citep{clinchant2016transductive,van2017unsupervised,liang2019distant} follow this learning mechanism and adapt the source classifier to unlabeled target features.
To gain benefits from end-to-end representation learning, researchers are more interested in generalization with deep models.
Such a setting without access to source data during adaptation is termed as source data-absent domain adaptation \citep{liang2020we,liang2021source}, model adaptation \citep{li2020model}, and source-free domain adaptation \citep{kundu2020universal}, respectively.
For the sake of simplicity, we utilize the term \emph{test-time domain adaptation} and give a unified definition.

\begin{definition}[Test-Time Domain Adaptation, TTDA]
Given a well-trained classifier $f_\mathcal{S}: \mathcal{X}_\mathcal{S} \to \mathcal{Y}_\mathcal{S}$ on the source domain $\mathcal{D}_\mathcal{S}$ and an unlabeled target domain $\mathcal{D}_\mathcal{T}$, \emph{test-time domain adaptation} aims to leverage the labeled knowledge implied in $f_\mathcal{S}$ to infer labels of all the samples in $\mathcal{D}_\mathcal{T}$, in a transductive learning \citep{joachims1999transductive} manner. Note that, all test data (target data) are required to be seen during adaptation.
\end{definition}

\setlength{\tabcolsep}{4.0pt}
\begin{table}[!t]
\caption{A taxonomy on TTDA methods with representative strategies.}
\resizebox{0.49\textwidth}{!}{
    \begin{tabular}{lll}
        \toprule
        \textbf{Families} & \textbf{Model Rationales} & \textbf{Representative Strategies}\\
        \midrule
        & \textbf{centroid-based} & SHOT \citep{liang2020we,liang2021source}, BMD \citep{qu2022bmd} \\
        \textbf{pseudo-} & \textbf{neighbor-based} & NRC \citep{yang2021exploiting}, SSNLL \citep{chen2022self}  \\
        \textbf{labeling} & \textbf{complementary labels} & LD \citep{you2021domain}, ATP \citep{wang2022source} \\
        & \textbf{optimization-based} & ASL \citep{yan2021augmented}, KUDA \citep{sun2022prior}  \\
        \midrule 
        & \textbf{data variations} & G-SFDA \citep{yang2021generalized}, APA \citep{sun2023domain} \\
        \textbf{consistency} & \textbf{model variations} & SFDA-UR \citep{sivaprasad2021uncertainty}, FMML \citep{peng2022toward}\\
        & \textbf{both variations} & AdaContrast \citep{chen2022contrastive}, MAPS \citep{ding2023maps} \\
        \midrule 
        & \textbf{entropy minimization} & ASFA \citep{xia2022privacy}, 3C-GAN \citep{li2020model}\\
        \textbf{clustering} & \textbf{mutual information} & SHOT \citep{liang2020we,liang2021source}, UMAD \citep{liang2021umad} \\
        & \textbf{explicit clustering} & ISFDA \citep{li2021imbalanced}, SDA-FAS \citep{liu2022source_eccv} \\
        \midrule 
        & \textbf{data generation} & 3C-GAN \citep{li2020model}, DI \citep{nayak2021mining} \\
        \textbf{source} & \textbf{data translation} & SFDA-IT \citep{hou2020source}, ProSFDA \citep{hu2022prosfda}\\
        \textbf{estimation} & \textbf{data selection} & SHOT++ \citep{liang2021source}, DaC \citep{zhang2022divide} \\
        & \textbf{feature estimation} & VDM-DA \citep{tian2022vdm}, CPGA \citep{qiu2021source}\\
        \midrule
        \textbf{self-supervision} & \textbf{auxiliary tasks} & SHOT++ \citep{liang2021source}, StickerDA \citep{kundu2022concurrent}\\
        \bottomrule
\end{tabular}}
\end{table}

\renewcommand{\arraystretch}{1.5}
\begin{figure}[tbp]
\centering
\begin{tabular}{c} 
 \includegraphics[trim={0.0cm 0.0cm 0.0cm 0.0cm},clip, width=0.4\textwidth]{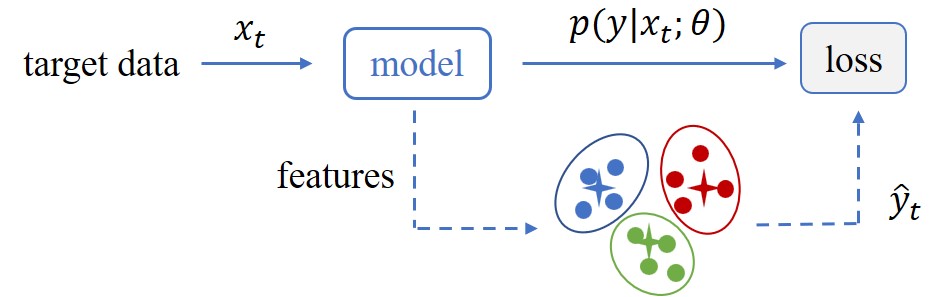}\\
 (a) centroid-based pseudo labels \\
 \includegraphics[trim={0.0cm 0.0cm 0.0cm 0.0cm},clip, width=0.4\textwidth]{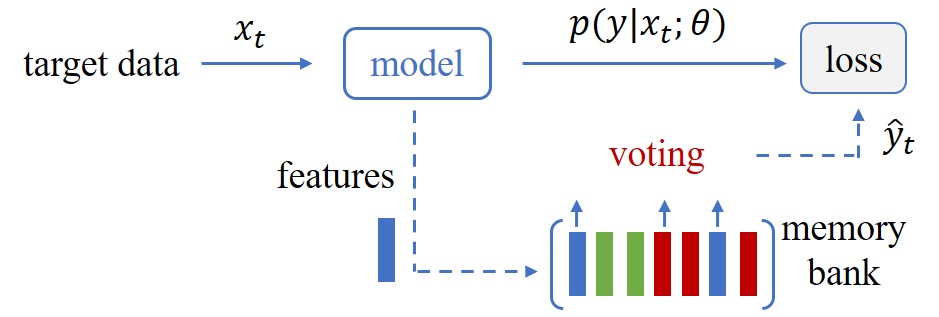}\\ 
 (b) neighbor-based pseudo labels \\
 \includegraphics[trim={0.0cm 0.0cm 0.0cm 0.0cm},clip, width=0.4\textwidth]{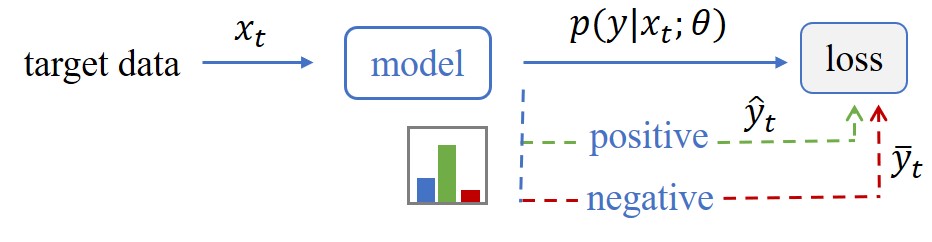}\\ 
 (c) complementary pseudo labels \\ 
\end{tabular}
\caption{\tim{
Three representative types of pseudo-labeling, where $\theta$ represents the model parameters, and $\hat{y}_t$ (or $\bar{y}_t$) denotes the pseudo label of the instance $x_t$.
}}
\label{fig:TTDA-PL}
\end{figure}

\subsection{Taxonomy on TTDA Algorithms}
\subsubsection{Pseudo-labeling}
To adapt a pre-trained model to an unlabeled target domain, a majority of TTDA methods take inspiration from the semi-supervised learning (SSL) field \citep{chen2022semi} and employ various prevalent SSL techniques tailored for unlabeled data during adaptation.
A simple yet effective technique, pseudo-labeling \citep{lee2013pseudo}, aims to assign a class label $\hat{y} \in \mathbb{R}^{C}$ for each unlabeled sample $x$ in $\mathcal{X}_t$ and optimize the following supervised learning objective to guide the learning process, 
\setlength\abovedisplayskip{0.2cm}
\setlength\belowdisplayskip{0.2cm}
\begin{equation}
    \min_\theta \mathbb{E}_{\{x,\hat{y}\} \in \mathcal{D}_t}\; w_{pl}(x)\cdot {d}_{pl}(\hat{y}, p(y|x;\theta)),
    \label{eq:pl}
\end{equation}
where $w_{pl}(x)$ denotes the real-valued weight associated with each pseudo-labeled sample $\{x, \hat{y}\}$, and $d_{pl}(\cdot)$ denotes the divergence between the predicted label probability distribution and the pseudo label probability $\hat{y}$, \eg, $-\sum_c \hat{y}_c \log [p(y|x;\theta)]_c$ if using the cross entropy as the divergence measure.
Since the pseudo labels of target data are inevitably inaccurate under domain shift, there exist three different solutions: (1) improving the quality of pseudo labels via denoising; (2) filtering out inaccurate pseudo labels with $w_{pl}(\cdot)$; and (3) developing a robust divergence measure ${d}_{pl}(\cdot,\cdot)$ for pseudo-labeling. 
To reduce the effects of noisy pseudo labels based on the argmax operation \citep{kim2021domain,li2021free,chen2021source}, most TTDA methods (\eg, SFIT \citep{hou2021visualizing}) consider only reliable pseudo labels using diverse filtering mechanisms.
\tim{Fig.~\ref{fig:TTDA-PL} illustrates three representative types of pseudo-labeling, which will be elaborated in the following part.}

\method{Centroid-based pseudo labels}
Inspired by a classic self-supervised approach, DeepCluster \citep{caron2018deep}, SHOT \citep{liang2020we,liang2021source} resorts to target-specific clustering for denoising the pseudo labels. 
The key idea is to obtain target-specific class centroids based on the network predictions and the target features and then derive the unbiased pseudo labels via the nearest centroid classifier. 
Formally, the class centroids and pseudo labels are updated as follows,
\setlength\abovedisplayskip{0.1cm}
\setlength\belowdisplayskip{0.1cm}
\begin{equation}
\renewcommand{\arraystretch}{1.1}
\left\{
\begin{array}{lr}
m_c = \sum_{x} [p_\theta(y_c|x)\cdot g(x)] / \sum_{x} p_\theta(y_c|x), \ c\in[1,C], &  \\
\hat{y} = \arg\min_c d(g(x), m_c),\ \forall x \in \mathcal{D}_t, &  
\end{array}
\right.
\label{eq:pl-cluster}
\end{equation}
where $p_\theta(y_c|x)=[p(y|x;\theta)]_c$ denotes the probability associated with the $c$-th class, and $g(x)$ denotes the feature of input $x$.
$m_c$ denotes the $c$-th class centroid, and $d(\cdot,\cdot)$ denotes the cosine distance function.
As class centroids always contain robust discriminative information and meanwhile weaken the category imbalance problem, this label refinery is prevalent in follow-up TTDA studies \citep{zhang2022divide,tang2021model,qiu2021source}.

Twofer \citep{liu2023twofer} identifies confident samples to build more accurate centroids, while BMD \citep{qu2022bmd} posits that a coarse centroid may not effectively represent ambiguous data and instead employs K-means clustering to discover multiple prototypes for each class.
Additionally, CoWA-JMDS \citep{lee2022confidence} performs Gaussian Mixture Modeling (GMM) in the target feature space to obtain the log-likelihood and pseudo label of each sample.
Apart from hard pseudo labels, FAUST \citep{lee2022feature} explores soft pseudo labels based on the class centroids, \eg, $[\hat{y}]_c = \frac{exp(-d(g(x), m_c)/\tau)}{\sum_c exp(-d(g(x), m_c)/\tau)}$, where $\tau$ denotes the temperature.
In contrast, BMD \citep{qu2022bmd} employs the exponential moving average (EMA) technique to dynamically accumulate the class centroids in mini-batches.

\method{Neighbor-based pseudo labels}
Another prevalent label denoising technique is to generate pseudo labels by incorporating the predictions of neighboring labels, relying on the assumption of local smoothness \citep{chen2022self,wang2022exploring,cao2021towards,chen2022contrastive,ding2022proxymix}.
For instance, SSNLL \citep{chen2022self} performs K-means clustering in the target domain and aggregates predictions of its neighbors within the same cluster.
DIPE \citep{wang2022exploring} diminishes label ambiguity by correcting the pseudo label to the majority vote of its neighbors.
In contrast, SFDA-APM \citep{kim2021domain} constructs an anchor set comprising only highly confident target samples and employs a point-to-set distance function to generate the pseudo labels. 
CAiDA \citep{dong2021confident} proposes a greedy chain-search strategy to find its nearest neighbor in the anchor set, interpolates its nearest anchor to the target feature, and uses the prediction of the synthetic feature instead. 

Inspired by neighborhood aggregation \citep{liang2021domain}, a few works \citep{cao2021towards,chen2022contrastive,ding2022proxymix,litrico2023guiding} maintain a memory bank storing both features and predictions of the target data $\{g(x_i),q_i\}_{i=1}^{n_t}$, allowing online refinement of pseudo labels.
Typically, the refined pseudo label is obtained through $\hat{p}_i = \frac{1}{m}\sum\nolimits_{j \in \mathcal{N}_i} q_j$, where $\mathcal{N}_i$ denotes the indices of $m$ nearest neighbors of $g(x_i)$ in the memory bank.
Specifically, ProxyMix \citep{ding2022proxymix} sharpens the network output $\bar{p}$ with the class frequency to avoid class imbalance, while NRC \citep{yang2021exploiting} devises a weighting scheme for neighbors during aggregation.
Instead of using the soft pseudo label $\hat{p}$, AdaContrast \citep{chen2022contrastive} utilizes the hard pseudo label with the argmax operation.

\method{Complementary pseudo labels}
Motivated by the idea of negative learning \citep{kim2019nlnl}, PR-SFDA \citep{luo2021exploiting} randomly chooses a label from the set $\{1, \dots, C\} \backslash \{\hat{y}_i\}$ as the complementary label $\bar{y}_i$ and thus optimizes the following loss function,
\setlength\abovedisplayskip{0.2cm}
\setlength\belowdisplayskip{0.2cm}
\begin{equation}
    \min_\theta - \sum\nolimits_{i=1}^{n_t} \sum\nolimits_{c=1}^{C} \mathds{1}(\bar{y}_i = c) \log (1 - p_\theta(y_c|x_i)),
    \label{eq:npl}
\end{equation}
where $\hat{y}_i$ denotes the inferred hard pseudo label. 
$\bar{y}$ is referred to as a negative pseudo label, indicating that the given input does not belong to this label. 
The probability of correctness is $\frac{C-1}{C}$ for the complementary label $\bar{y}_i$, providing correct information even from wrong labels $\hat{y}_i$.
LD \citep{you2021domain} develops a heuristic strategy to randomly select an informative complementary label with medium prediction scores.
Besides, NEL \citep{ahmed2022cleaning} and PLUE \citep{litrico2023guiding} randomly select multiple complementary labels, except for the inferred pseudo label, and optimizes the multi-class variant of Eq.~(\ref{eq:npl}).
ATP \citep{wang2022source} further generates multiple complementary labels according to a pre-defined threshold on prediction scores.

\method{Optimization-based pseudo labels}
By leveraging the prior knowledge of the target label distribution like class balance \citep{zou2018unsupervised}, some TTDA methods \citep{you2021domain,sivaprasad2021uncertainty,huang2021model} vary the threshold for each class so that a certain proportion of points per class are selected.
Such a strategy helps avoid the `winner-takes-all' dilemma where the pseudo labels come from several major categories, potentially deteriorating the following training process.
Furthermore, ASL \citep{yan2021augmented} directly imposes the equi-partition constraint on the pseudo labels $\hat{p}_{i}$ and solves the optimization problem below,
\setlength\abovedisplayskip{0.2cm}
\setlength\belowdisplayskip{0.2cm}
\begin{equation}
\begin{aligned}
    \min_{\hat{p}_{i}} & - \sum\nolimits_i\sum\nolimits_c \hat{p}_{ic} \log p_\theta(y_c|x_i) + \lambda \sum\nolimits_i\sum\nolimits_c \hat{p}_{ic} \log \hat{p}_{ic}, \\
    s.t. \; & \forall i,c:\; \hat{p}_{ic}\in [0,1], \;\sum\nolimits_c \hat{p}_{ic}=1, \;\sum\nolimits_i \hat{p}_{ic}=\frac{n_t}{C}.
\end{aligned}
\end{equation}
Likewise, IterNLL \citep{zhang2021unsupervised} provides a closed-form solution of $\{\hat{p}\}$ under the uniform prior assumption.
KUDA \citep{sun2022prior} even introduces a hard constraint $\hat{p}_{ic}\in \{0,1\}$ and solves the zero-one programming problem.
\tim{In addition, ReCLIP \cite{hu2024reclip} constructs the affinity graph and employs label propagation to produce closed-form pseudo labels.}

\method{Ensemble-based pseudo labels}
Rather than relying on a single noisy pseudo label, ISFDA \citep{li2021imbalanced} generates a secondary pseudo label to aid the primary one.
Besides, ASL \citep{yan2021augmented} and C-SFDA \citep{karim2023csfda} adopt a weighted average of predictions under multiple random data augmentation, while ELR \citep{yi2023when} ensembles historical predictions from previous training epochs.
NEL \citep{ahmed2022cleaning} further aggregates logits under different data augmentation and trained models simultaneously.
Inspired by a classic semi-supervised learning method \citep{laine2017temporal}, some TTDA methods \citep{liang2022dine,panagiotakopoulos2022online} maintain an EMA of predictions at different time steps as pseudo labels.
Moreover, C-SFDA \citep{karim2023csfda} maintains a mean teacher model \citep{tarvainen2017mean} that generates pseudo labels for the current student network.
Additionally, other methods attempt to generate pseudo labels based on predictions from various models, \eg, multiple source models \citep{liang2022dine,li2022union}, a multi-head classifier \citep{kundu2021generalize}, and models from both domains \citep{hou2021visualizing}.
In particular, SFDA-VS \citep{ye2021source} follows MC dropout \citep{gal2016dropout} and obtains the final prediction through multiple forward passes.

Another line of ensemble-based TTDA methods \citep{cao2021towards,yan2021augmented,xiong2022source} aims to integrate predictions from different labeling criteria using a weighted average. 
For example, e-SHOT-CE \citep{cao2021towards} utilizes both centroid-based and neighbor-based pseudo labels.
Besides the weighting scheme, other approaches \citep{qiu2021source,dong2021confident,wang2022exploring,kumar2023conmix} explore different labeling criteria in a cascade manner.
For instance, DIPE \citep{wang2022exploring} employs the neighbor-based labeling criterion with centroid-based pseudo labels.

\method{Learning with pseudo labels}
Existing pseudo-labeling-based TTDA methods have employed various robust divergence measures ${d}_{pl}$.
Generally, most methods utilize the standard cross-entropy loss for all target samples with hard pseudo labels \citep{liang2020we,yan2021augmented} or soft pseudo labels \citep{tang2021model,deng2021universal}.
Note that several methods \citep{ding2022proxymix,tian2023source} convert hard pseudo labels into soft pseudo labels using the label smoothing trick \citep{muller2019does}.
As pseudo labels are noisy, many TTDA methods incorporate an instance-specific weighting scheme into the standard cross-entropy loss, including hard weights \citep{kim2021domain,hou2021visualizing,chen2021source}, and soft weights \citep{huang2021model,ye2021source}.
Besides, AUGCO \citep{prabhu2022augco} considers the class-specific weight in the cross-entropy loss to mitigate label imbalance.
In addition to the cross-entropy loss, alternative choices include the generalized cross entropy \citep{rusak2022if}, the inner product distance between the pseudo label and the prediction \citep{yang2021exploiting,qiu2021source}, and a new discrepancy measure $\log(1-\hat{y}^Tp(y|x;\theta))$ \citep{yi2023when}.
Moreover, BMD \citep{qu2022bmd} and OnDA \citep{panagiotakopoulos2022online} employ the symmetric cross-entropy loss to guide the self-labeling process.
CATTAn \citep{thopalli2022domain} exploits the negative log-likelihood ratio between correct and competing classes.

\renewcommand{\arraystretch}{1.5}
\begin{figure}[tbp]
\centering
\begin{tabular}{c} 
 \includegraphics[trim={0.0cm 0.0cm 0.0cm 0.0cm},clip, width=0.4\textwidth]{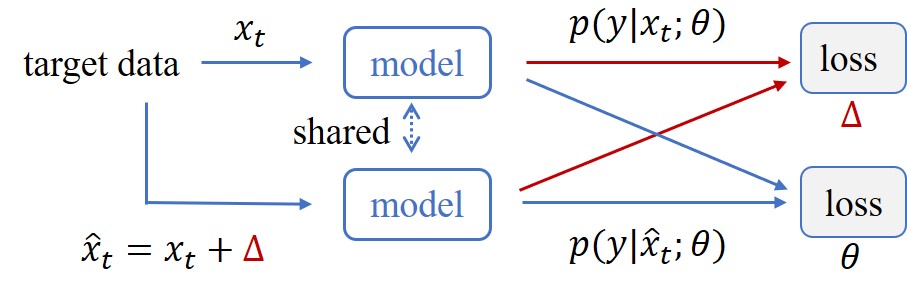}\\
 (a) consistency under data variations \\
 \includegraphics[trim={0.0cm 0.0cm 0.0cm 0.0cm},clip, width=0.4\textwidth]{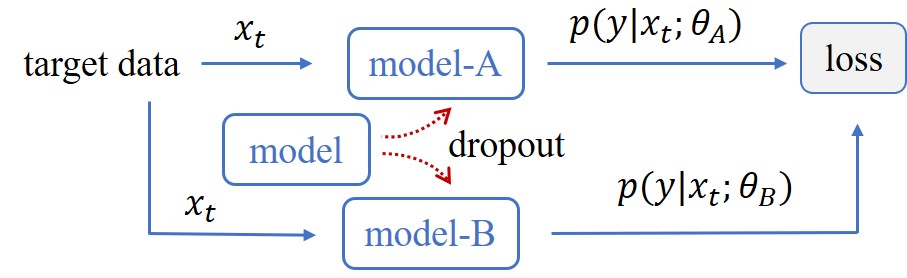}\\ 
 (b) consistency under model variations \\
 \includegraphics[trim={0.0cm 0.0cm 0.0cm 0.0cm},clip, width=0.4\textwidth]{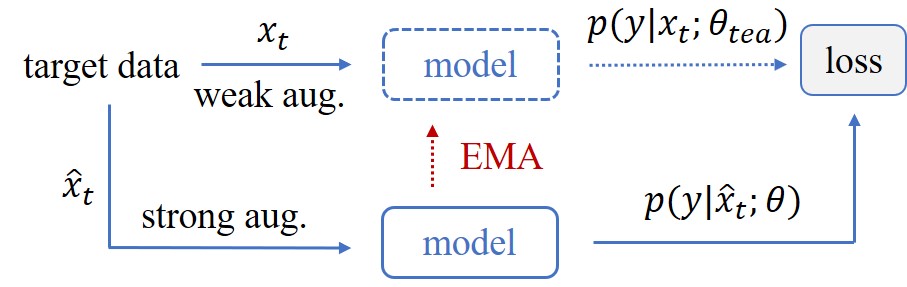}\\ 
 (c) consistency under data \& model variations \\ 
\end{tabular}
\caption{\tim{
Three representative types of consistency training, where $\hat{x}_t$ represents the data variant of $x_t$, and $\theta_A$ (or $\theta_B$ and $\theta_{tea}$) denotes the model variant of $\theta$.
}}
\label{fig:TTDA-CT}
\end{figure}

\subsubsection{Consistency Training}
Consistency regularization, a prevailing strategy in recent semi-supervised learning literature \citep{yang2021survey,chen2022semi}, is primarily built on the smoothness assumption or the manifold assumption.
It aims to enforce consistent network predictions or features under variations in the input data space or the model parameter space.
Moreover, another line of consistency training methods attempts to match the statistics of different domains even without the source data. 
\tim{Fig.~\ref{fig:TTDA-CT} illustrates three representative types of consistency training, which will be elaborated in the following part.}

\method{Consistency under data variations}
Benefiting from advanced data augmentation techniques such as RandAugment \citep{cubuk2020randaugment}, several prominent semi-supervised learning methods \citep{xie2020unsupervised,sohn2020fixmatch} unleash the power of consistency regularization over unlabeled data that can be effortlessly adopted in TTDA approaches.
An exemplar of consistency regularization \citep{sohn2020fixmatch} is expressed as:
\setlength\abovedisplayskip{0.2cm}
\setlength\belowdisplayskip{0.2cm}
\begin{equation}
    \mathcal{L}_{fm}^{con} = \frac{1}{n_t}\sum_{i=1}^{n_t} \text{CE}\left(p_{\tilde{\theta}}(y|x_i), p_{\theta}(y|\hat{x}_i)\right),
\label{eq:fixmatch}
\end{equation}
where $p_{\theta}(y|x_i)=p(y|x_i;\theta)$, and $\text{CE}(\cdot,\cdot)$ refers to cross-entropy between two distributions.
Besides, $\hat{x}_i$ represents the variant of $x_i$ under another augmentation transformation, and $\tilde{\theta}$ is a fixed copy of current network parameters $\theta$.
Another representative consistency regularization is virtual adversarial training (VAT) \citep{miyato2018virtual}, which devises a smoothness constraint as follows,
\setlength\abovedisplayskip{0.2cm}
\setlength\belowdisplayskip{0.2cm}
\begin{equation}
    \mathcal{L}_{vat}^{con} = \frac{1}{n_t}\sum_{i=1}^{n_t} \max_{\|\Delta_i\| \leq \epsilon}[\text{KL}(p_{\tilde{\theta}}(y|x_i) \;||
    \; p_{\theta}(y|x_i + \Delta_i))],
\label{eq:vat}
\end{equation}
where $\Delta_i$ is a perturbation that disperses the prediction most within an intensity range of $\epsilon$ for the target data $x_i$, and $\text{KL}$ denotes the Kullback–Leibler divergence.

ATP \citep{wang2022source} directly employs the same consistency regularization in Eq.~(\ref{eq:fixmatch}), while other TTDA methods \citep{wang2021learning,chen2022contrastive,zhang2022divide,kumar2023conmix} replace $p_{\tilde{\theta}}(y|x_i)$ with hard pseudo labels for target data under weak augmentation, followed by a cross-entropy loss for target data under strong augmentation.
Note that, many of these hard labels are obtained using the label denoising techniques mentioned earlier.
Apart from strong augmentations, ProSFDA \citep{hu2022prosfda} and SFDA-FSM \citep{yang2022source} require learning the domain translation module first, and ProSFDA seeks feature-level consistency under different augmentations at the same time.
TeST \citep{sinha2023test} introduces a flexible mapping network to match features under two different augmentations. 
On the contrary, OSHT \citep{feng2021open} maximizes the mutual information between the predictions of two different transformed inputs to retain the semantic information as much as possible.

Following the objective in Eq.~(\ref{eq:vat}), another line of TTDA methods \citep{li2020model,yan2021augmented} attempts to encourage consistency between target samples with their data-level neighbors, while APA \citep{sun2023domain} learns the neighbors in the feature space.
Instead of generating the most divergent neighbor $x_i+\Delta_i$ according to the predictions, JN \citep{li2022jacobian} devises a Jacobian norm regularization to control the smoothness in the neighborhood of the target sample.
Furthermore, G-SFDA \citep{yang2021generalized} discovers multiple neighbors from a memory bank and minimizes their inner product distances over the predictions.
Moreover, Mixup \citep{zhang2018mixup} performs linear interpolations on two inputs and their corresponding labels, which can be treated as seeking consistency under data variation \citep{liang2022dine,lee2022confidence,kumar2023conmix}.

\method{Consistency under model variations}
Reducing model uncertainty \citep{gal2016dropout} is also beneficial for learning robust features for TTDA tasks, on top of uncertainty measured with input change.
Following MC dropout \citep{gal2016dropout}, FAUST \citep{lee2022feature} activates dropout in the model and performs multiple stochastic forward passes to estimate the epistemic uncertainty.
SFDA-UR \citep{sivaprasad2021uncertainty} appends multiple extra dropout layers behind the feature encoder and minimizes the mean squared error (MSE) between predictions as uncertainty.
Further, ASFA \citep{xia2022privacy} adds different perturbations to the intermediate features to promote predictive consistency.
FMML \citep{peng2022toward} offers another form of model variation by network slimming and sought predictive consistency across different networks.

Another consistency regularization requires the existence of both the source and target models and thus minimizes the difference across different models, such as feature-level discrepancy \citep{kothandaraman2023salad} and output-level discrepancy \citep{liang2022dine,conti2022cluster,sinha2023test}.
Furthermore, the mean teacher framework \citep{tarvainen2017mean} is also utilized to form a strong teacher model and a learnable student model. 
The teacher and the student models share the same architecture, and the weights of the teacher model $\theta_{tea}$ are gradually updated by $\theta_{tea} = (1-\eta) \theta_{tea} + \eta \theta$, where $\theta$ denotes the weights of the student model, and $\eta$ is the momentum coefficient.
Therefore, the mean teacher model is regarded as a temporal ensemble of student models with more accurate predictions.
In reality, a few TTDA methods including \citep{lao2021hypothesis} consider the multi-head classifier and promote consistent predictions by different heads. 

\method{Consistency under data \& model variations}
In reality, data variation and model variation could be integrated into a unified framework.
For example, the mean teacher framework \citep{tarvainen2017mean} is enhanced by blending strong data augmentation techniques, and the discrepancy between predictions of the student and teacher models is minimized as follows, 
\setlength\abovedisplayskip{0.2cm}
\setlength\belowdisplayskip{0.2cm}
\begin{equation}
    \mathcal{L}_{mt}^{con} = \mathbb{E}_{x \in \mathcal{D}_t} d_{mt}(p(y|x, \theta), p(y|\tau(x), \theta_{tea})),
\end{equation}
where $\tau(\cdot)$ denotes the strong data augmentation, and $d_{mt}$ denotes the divergence measure, \eg, the KL divergence \citep{liu2022source_eccv,hou2021visualizing}, the MSE loss \citep{zhang2021source}, and the cross-entropy loss \citep{chen2022self}.
Besides, several methods \citep{vs2022target,liu2022source,huang2021model,li2022source_cvpr} attempt to extract useful information from the teacher and employ task-specific loss functions to seek consistency.
Apart from the output-level consistency, TT-SFUDA \citep{vs2022target} matches the features extracted by different models with the MSE distance, while AdaContrast \citep{chen2022contrastive} and PLUE \citep{litrico2023guiding} learn semantically consistent features like MoCo \citep{he2020momentum}.

Instead of strong data augmentations, LODS \citep{li2022source_cvpr} and SFIT \citep{hou2021visualizing} use the style transferred image instead, MAPS \citep{ding2023maps} considers spatial transforms, and SMT \citep{zhang2021source} elaborates the domain-specific perturbation by averaging the target images. 
Different from model variations in the mean teacher scheme, OnTA \citep{wang2021ontarget} distills knowledge from the source model to the target model, while HCL \citep{huang2021model} promotes feature-level consistency among the current model and historical model.

\method{Miscellaneous consistency regularizations}
To prevent excessive deviation from the original source model, a flexible strategy is adopted by a few TTDA methods \citep{li2020model,xiong2022source} by establishing a parameter-based regularization term $\|\theta_s - \theta\|_2^2$, where $\theta_s$ is the fixed source weight.
Another line of research focuses on matching the batch normalization (BN) statistics (\ie, the mean and the variance), across models with different measures, such as the KL divergence \citep{ishii2021source} and the MSE error \citep{zhang2021source,ahmed2022cross}, whereas OSUDA \citep{liu2021adapting} encourages the learned scaling and shifting parameters in BN layers to be consistent.
Similarly, an explicit feature-level regularization \citep{liu2021ttt++} is devised to match the first and second-order moments of features in different domains. 

As for the network architecture in the target domain, a unique design termed dual-classifier is utilized to seek robust domain-invariant representations.
For example, BAIT \citep{yang2021casting} introduces an extra $C$-dimensional classifier to the source model, forming a dual-classifier model with a shared feature encoder. 
During adaptation in the target domain, the shared feature encoder and the new classifier are trained with the classifier from the source domain head fixed. 
Such a training scheme has also been utilized by many TTDA methods \citep{tian2023source,wang2022exploring,xia2021adaptive,sivaprasad2021uncertainty,xia2022privacy} through modeling the consistency between different classifiers.
Besides, SFDA-APM \citep{kim2021domain} develops a self-training framework that optimizes the shared feature encoder and two classification heads with different pseudo-labeling losses, respectively. 

\renewcommand{\arraystretch}{1.5}
\begin{figure}[tbp]
\centering
\begin{tabular}{c} 
 \includegraphics[trim={0.0cm 0.0cm 0.0cm 0.0cm},clip, width=0.4\textwidth]{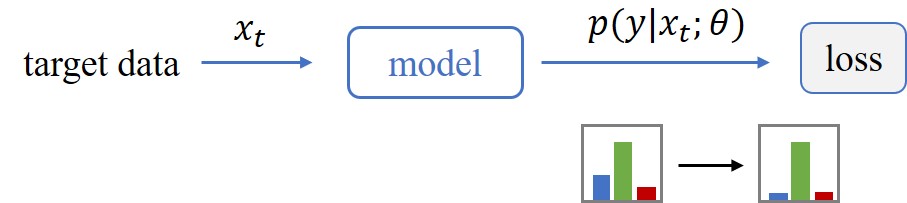}\\
 (a) uncertainty minimization over network predictions \\
 \includegraphics[trim={0.0cm 0.0cm 0.0cm 0.0cm},clip, width=0.4\textwidth]{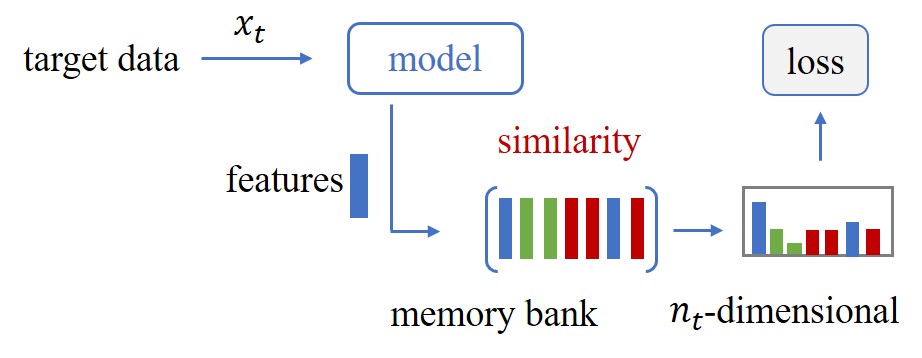}\\ 
 (b) clustering-promoting over network features \\
\end{tabular}
\caption{\tim{
Two representative types of clustering-based training, where similarity is obtained based on a feature memory bank.
}}
\label{fig:TTDA-C}
\end{figure}

\subsubsection{Clustering-based Training}
Except for the pseudo-labeling paradigm, nearly all semi-supervised learning algorithms rely on the cluster assumption \citep{yang2021survey}, which asserts that the decision boundary should not cross high-density regions, but instead lie in low-density regions. 
As a result, another popular category of TTDA approaches favors low-density separation by reducing the uncertainty of the target network predictions \citep{liang2020we,li2020model} or promoting clustering among the target features \citep{li2021imbalanced,qiu2021source}.
\tim{Fig.~\ref{fig:TTDA-C} illustrates these two representative types of clustering-based training, which will be elaborated in the following part.}

\method{Entropy minimization} ASFA \citep{xia2022privacy} utilizes robust measures from information theory to encourage confident predictions for unlabeled target data. To achieve this, it minimizes the $\alpha$-Tsallis entropy given by:
\setlength\abovedisplayskip{0.2cm}
\setlength\belowdisplayskip{0.2cm}
\begin{equation}
    \mathcal{L}_{tsa} = \frac{1}{n_t} \sum_{i=1}^{n_t} \frac{1}{\alpha - 1} [1 - \sum_{c=1}^{C} p_\theta(y_c|x_i)^\alpha],
\label{eq:tsa}
\end{equation}
where $\alpha>$ 0 is called the entropic index. 
Note that, as $\alpha$ approaches 1, the Tsallis entropy converges to the standard Shannon entropy, given by $\mathcal{H}(p_\theta(y|x_i)) = \sum_c p_\theta(y_c|x_i) \log p_\theta(y_c|x_i)$.
In practice, the conditional Shannon entropy $\mathcal{H}(p_\theta(y|x))$ has been widely used in TTDA methods \citep{li2020model,liu2021adapting,sivaprasad2021uncertainty,you2021domain,kundu2021generalize,bateson2022source,sinha2023test}.
Besides, there exist numerous variations of standard entropy minimization.
For instance, SFDA-VS \citep{ye2021source} develops a nonlinear weighted entropy minimization loss that emphasizes low-entropy samples.
TT-SFUDA \citep{vs2022target} focuses on the entropy of the ensemble predictions under multiple augmentations.

When $\alpha$ is set to 2, the Tsallis entropy in Eq.~(\ref{eq:tsa}) is equivalent to the maximum squares loss \citep{chen2019domain,liu2021source,kumar2023conmix}, given by $\sum_c p_\theta(y_c|x_i)^2$.
Compared to the Shannon entropy, the gradient of the maximum squares loss increases linearly, preventing easy samples from dominating the training process in the high probability region.
Building on this, Batch Nuclear-norm Maximization (BNM) \citep{cui2020towards} approximates the prediction diversity using the matrix rank, which is utilized by CDL \citep{wang2021learning}. 
Additionally, SI-SFDA \citep{ye2022alleviating} pays attention to the class confusion matrix and minimizes the inter-class confusion to ensure that no samples are ambiguously classified into two classes at the same time.

\method{Mutual information maximization}
Another favorable clustering-based regularization is mutual information maximization, which aims to maximize the mutual information \citep{shi2012information} between the inputs and the discrete labels as follows,
\setlength\abovedisplayskip{0.1cm}
\setlength\belowdisplayskip{0.1cm}
\begin{equation}
\begin{aligned}
    &\max_\theta \mathcal{I} (\mathcal{X}_t, \hat{\mathcal{Y}_t}) = \mathcal{H}(\hat{\mathcal{Y}_t}) - \mathcal{H}(\hat{\mathcal{Y}_t}|\mathcal{X}_t) = \\
    &-\sum_{c=1}^{C} \bar{p}_\theta(y_c) \log \bar{p}_\theta(y_c) + \frac{1}{n_t} \sum_{i=1}^{n_t}\sum_{c=1}^{C} p_\theta(y_c|x_i) \log p_\theta(y_c|x_i),
\end{aligned}
\label{eq:mi}
\end{equation}
where $\bar{p}_\theta(y_c)=\frac{1}{n_t}\sum_i p_\theta(y_c|x_i)$ denotes the $c$-th element in the estimated class label distribution.
Intuitively, increasing the extra diversity term $\mathcal{H}(\hat{\mathcal{Y}_t})$ promotes uniform distribution of target labels, circumventing the degenerate solution where each sample is assigned to the same class.
Such a regularization is initially introduced in SHOT \citep{liang2020we} and SHOT++ \citep{liang2021source} for image classification and then employed in plenty of TTDA methods \citep{ishii2021source,lao2021hypothesis,wang2021ontarget,li2022jacobian,wang2022exploring,litrico2023guiding}.
Instead of using the network prediction $p_\theta(y|x)$, GKD \citep{tang2021model} employs the ensemble prediction based on its neighbors for mutual information maximization.
DaC \citep{zhang2022divide} and U-SFAN \citep{roy2022uncertainty} introduce a balancing parameter between two terms in Eq.~(\ref{eq:mi}) to increase flexibility.
In particular, U-SFAN \citep{roy2022uncertainty} develops an uncertainty-guided entropy minimization loss by emphasizing low-entropy predictions, whereas ATP \citep{wang2022source} encompasses the instance-wise uncertainty in both terms of Eq.~(\ref{eq:mi}). 
VMP \citep{jing2022variational} further provides a probabilistic framework based on Bayesian neural networks and integrates mutual information into the likelihood function.

It is worth noting that the diversity term can be rewritten as $\mathcal{H}(\hat{\mathcal{Y}_t}) = -\text{KL}(\bar{p}_\theta(y)||\mathcal{U}) + \log C$, where $\bar{p}_\theta(y)$ denotes the average label distribution in the target domain, and $\mathcal{U}$ is a $C$-dimensional uniform vector.
This term alone has also been employed in numerous TTDA methods \citep{hou2021visualizing,yang2021exploiting,chen2022contrastive,kundu2022concurrent,panagiotakopoulos2022online,tian2023source,panagiotakopoulos2022online,thopalli2022domain} to prevent class collapse.
To better guide the learning process, a few works \citep{krause2010discriminative,hu2017learning} modify the mutual information regularization by substituting a reference class-ratio distribution in place of $\mathcal{U}$.
Unlike AdaMI \citep{bateson2022source}, which leverages the target class ratio as a prior, UMAD \citep{liang2021umad} utilizes the flattened label distribution within a mini-batch instead to mitigate the class imbalance problem, and AUGCO \citep{prabhu2022augco} maintains the moving average of the predictions as the reference distribution.

\renewcommand{\arraystretch}{1.5}
\begin{figure}[tbp]
\centering
\begin{tabular}{c} 
 \includegraphics[trim={0.0cm 0.0cm 0.0cm 0.0cm},clip, width=0.4\textwidth]{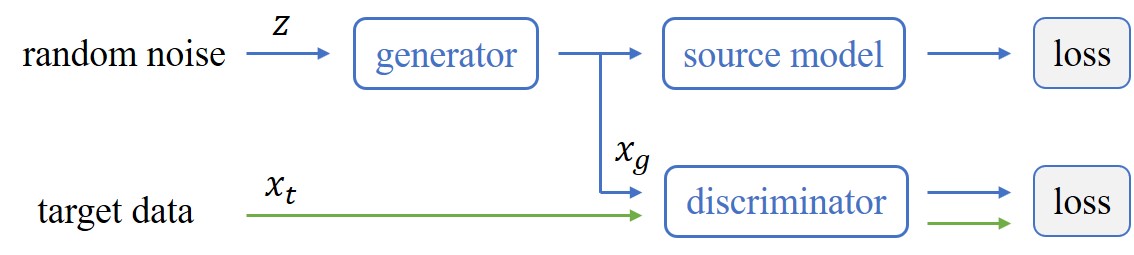}\\
 (a) data generation \\
 \includegraphics[trim={0.0cm 0.0cm 0.0cm 0.0cm},clip, width=0.4\textwidth]{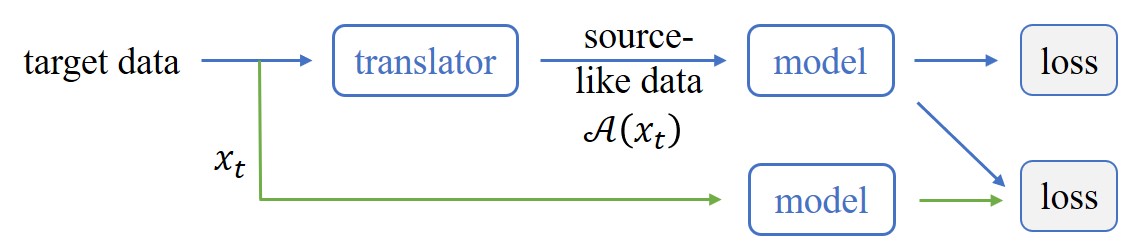}\\ 
 (b) data translation \\
 \includegraphics[trim={0.0cm 0.0cm 0.0cm 0.0cm},clip, width=0.4\textwidth]{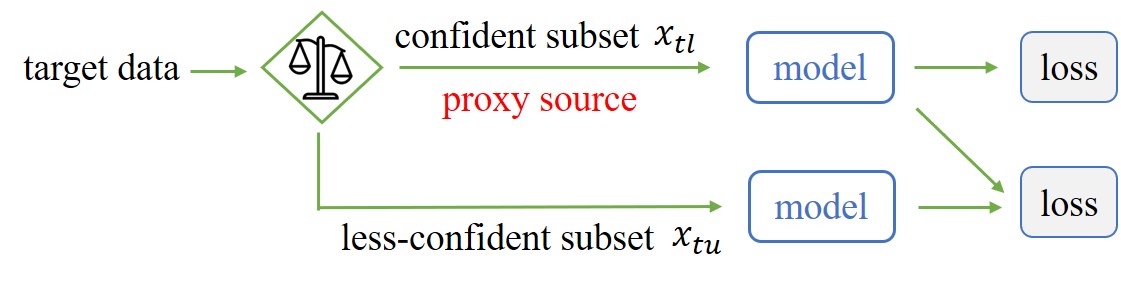}\\ 
 (c) data selection \\
\end{tabular}
\caption{\tim{
Three representative types of source distribution estimation, where surrogate source data is obtained through generation, translation, and selection, respectively.
}}
\label{fig:TTDA-SD}
\end{figure}

\subsubsection{Source Distribution Estimation}
Another favored family of TTDA approaches compensates for the absence of source data by inferring data from the pre-trained model, transforming the challenging TTDA problem into a well-studied DA problem.
Existing source estimation approaches could be categorized into three groups: data generation from random noises \citep{morerio2020generative,li2020model,kurmi2021domain}, data translation \citep{hou2021visualizing,yan2021source,zhou2022generative}, and data selection \citep{liang2021source,yang2021casting,ding2022proxymix}.
\tim{Fig.~\ref{fig:TTDA-SD} illustrates three representative types of source distribution estimation, which will be elaborated in the following part.}

\method{Data generation}
To generate valid target-style source samples, 3C-GAN \citep{li2020model} introduces a data generator $G(\cdot;\theta_G)$ conditioned on randomly sampled labels, along with a binary discriminator $D(\cdot;\theta_D)$.
The optimization objective is similar to the conditional GAN \citep{mirza2014conditional} that is written as follows:
\setlength\abovedisplayskip{0.2cm}
\setlength\belowdisplayskip{0.2cm}
\begin{equation}
\begin{aligned}
    \min_{\theta_G}&\max_{\theta_D}\ \mathbb{E}_{x_t \in \mathcal{X}_t}[\log D(x_t)] + \mathbb{E}_{y_t,z} [\log(1 - D(G(y_t,z)))] \\
    & - \lambda_s \mathbb{E}_{y_t,z} \sum\nolimits_{c}\mathds{1}(y_t=c) \log p(y_c|G(y_t,z),\theta),
\end{aligned}
\label{eq:cgan}
\end{equation}
where $z$ is a random noise vector, $y_t$ is a pre-defined label, $\lambda_s>0$ is a balancing parameter, and $\theta$ denotes the parameters of the pre-trained prediction model.
By alternately optimizing $\theta_G$ and $\theta_D$, the resulting class conditional generator $G$ can generate multiple surrogate labeled source instances for the subsequent domain alignment step, \ie, $\mathcal{D}_g = \{x_i,y_i\}_{i=1}^{n_g}$, where $x_i=G(y_i,z)$ and $n_g$ is the number of generated samples. 
PLR \citep{morerio2020generative} disregards the last term in Eq.~(\ref{eq:cgan}) to infer diverse target-like samples.
On the other hand, SDDA \citep{kurmi2021domain} maximizes the log-likelihood of generated data $x_g$ and employs two different domain discriminators, \ie, a data-level GAN discriminator and a feature-level domain discriminator.

In addition to adversarial training, DI \citep{nayak2021mining} performs Dirichlet modeling with the source class similarity matrix and then optimizes the noisy input to match its output with the sampled softmax vector $q$ as,
\setlength\abovedisplayskip{0.2cm}
\setlength\belowdisplayskip{0.2cm}
\begin{equation}
    x_g = \arg\min_x \text{CE}(q, p_\theta(y|x))
\end{equation}
which is referred to as data impression of the source domain.
Besides, SPGM \citep{yang2022revealing} first estimates the target distribution using GMM and then constrains the generated data to be derived from the target distribution. 

Motivated by recent advances in data-free knowledge distillation \citep{yin2020dreaming,liu2021data}, SFDA-KTMA \citep{liu2021source} exploits the moving average statistics of activations stored in BN layers of the pre-trained source model and imposes the following BN matching constraint on the generator, 
\setlength\abovedisplayskip{0.2cm}
\setlength\belowdisplayskip{0.2cm}
\begin{equation}
    \mathcal{L}_{bn} = \sum\nolimits_l \sum\nolimits_i \|\mu_{g,l}^{(i)} - \mu_{s,l}^{(i)}\|_2 + \|{\delta_{g,l}^{(i)}}^2 - {\delta_{s,l}^{(i)}}^2\|_2,
\label{eq:bns}
\end{equation}
where $B$ is the size of a mini-batch, $\mu_{s,l}^{(i)}$ and ${\delta_{s,l}^{(i)}}^2$ represent the corresponding running mean and variance stored in the source model, and $\mu_{g,l}^{(i)} = \frac{1}{B}\sum\nolimits_z f_l^{(i)}(x_g)$ and ${\delta_{g,l}^{(i)}}^2 = \frac{1}{B}\sum\nolimits_z (f_l^{(i)}(x_g) - \mu_{g,l}^{(i)})^2$ denote the batch-wise mean and variance estimates of the $i$-th feature channel at the $l$-th layer for synthetic data from the generator, respectively.
As indicated in \citep{li2017demystifying}, matching the BN statistics can aid in ensuring that the generated data resembles the source style.
SFDA-FSM \citep{yang2022source} further minimizes the $L_2$-norm difference between intermediate features (\aka, the content loss \citep{gatys2016image}) to preserve the content knowledge of the target domain.

\method{Data translation}
SSFT-SSD \citep{yan2021source} initializes $x_g$ as $x_t \in \mathcal{X}_t$ and directly performs optimization on the input space with the gradient of the $L_2$-norm regularized cross-entropy loss being zero.
On the contrary, SFDA-TN \citep{sahoo2020unsupervised} optimizes a learnable data transformation network that maps target data to the source domain such that the maximum class probability is maximized.
Inspired by the success of visual prompts \citep{bahng2022visual}, ProSFDA \citep{hu2022prosfda} adds a learnable image perturbation to all target data, enabling the BN statistics to be aligned with those stored in the source model.
Besides, the style-transferred image is obtained using spectrum mixup \citep{yang2020fda} between the target image and its perturbed image.

Another line of data translation methods \citep{hou2020source,hou2021visualizing,zhou2022generative} explicitly introduces an additional module $\mathcal{A}$ to transfer target data to source-like style.
In particular, SFDA-IT \citep{hou2020source} optimizes the translator with the style matching loss in Eq.~(\ref{eq:bns}) as well as the feature-level content loss, with the source model frozen.
Furthermore, SFDA-IT \citep{hou2020source} employs entropy minimization over the fixed source model to promote semantic consistency.
To improve the performance of style transfer, SFIT \citep{hou2021visualizing} further develops a variant of the style reconstruction loss \citep{gatys2016image} as follows,
\setlength\abovedisplayskip{0.2cm}
\setlength\belowdisplayskip{0.2cm}
\begin{equation}
    \mathcal{L}_{style} = \|g(x)g(x)^{T} - g(\mathcal{A}(x))g(\mathcal{A}(x))^{T}\|_2,
\end{equation}
where $g(x) \in \mathcal{R}^{n_c \times HW}$ denotes the reshaped feature map, and $H, W$ and $n_c$ represent the feature map height, width, and the number of channels, respectively. 
The channel-wise self correlations $g(x)g(x)^T$ are also known as the Gram matrix.
Additionally, SFIT \citep{hou2021visualizing} maintains the relationship of outputs between different networks.
GDA \citep{zhou2022generative} also relies on BN-based style matching and entropy minimization but further enforces the phase consistency and the feature-level consistency between the original image and the stylized image to preserve the semantic content.

\method{Data selection}
In addition to synthesizing source samples through data generation or data translation, another family of TTDA methods \citep{liang2021source,yang2021casting,ding2022proxymix,wang2022source,chen2022self,yang2023divide} selects source-like samples from the target domain as surrogate source data, greatly reducing computational costs.
Typically, the whole target domain is divided into two splits, \ie, a labeled subset $\hat{\mathcal{X}}_{tl}$ and an unlabeled subset $\hat{\mathcal{X}}_{tu}$, where the labeled subset acts as the inaccessible source domain.
Based on the network outputs of the adapted model in the target domain, SHOT++ \citep{liang2021source} makes the first attempt towards data selection by selecting low-entropy samples in each class for an extra intra-domain alignment step.
Such an adapt-and-divide strategy has been adopted in later works \citep{ye2021source,liu2021source,wang2022source} where the ratio or the number of selected samples per class is always kept same to prevent severe class imbalance.
DaC \citep{zhang2022divide} utilizes the maximum softmax probability instead of the entropy criterion.
Furthermore, BETA \citep{yang2023divide} constructs a two-component GMM over all the target features to separate the confident subset $\hat{\mathcal{X}}_{tl}$ from the less confident subset $\hat{\mathcal{X}}_{tu}$.

Apart from the adapted target model, a few approaches \citep{ding2022proxymix,liu2022self,huang2022relative} utilize the source model to partition the target domain before the intra-domain adaptation step. 
For each class separately, MTRAN \citep{huang2022relative} selects the low-entropy sample, ProxyMix \citep{ding2022proxymix} leverages the distance from target features to source class prototypes, and SAB \citep{liu2022self} adopt the maximum prediction probability.
To simulate the source domain more accurately, MTRAN \citep{huang2022relative} further applies the mixup augmentation technique after the dataset partition step.
On the other hand, some TTDA methods \citep{yang2021casting,xia2021adaptive,ye2021source,chen2022self,chu2022denoised,tian2023source} do not fix the domain partition but alternately update the domain partition and learn the target model in the adaptation step.
For instance, SSNLL \citep{chen2022self} follows the small loss trick for noisy label learning and assigns samples with small loss to the labeled subset at the beginning of each epoch.
On top of the global division, BAIT \citep{yang2021casting} and SFDA-KTMA \citep{liu2021source} split each mini-batch into two sets based on the criterion of entropy ranking, while D-MCD \citep{chu2022denoised} employs the classifier determinacy disparity and the agreement between different self-labeling strategies.

\method{Feature estimation}
In contrast to source data synthesis, previous works \citep{qiu2021source,tian2022vdm,ding2022source} provide a cost-effective alternative by simulating the source features.
MAS$^3$ \citep{stan2021unsupervised} and LDAuCID \citep{rostami2021lifelong} require learning a GMM over the source features before model adaptation, which may not hold in real-world scenarios.
Instead, VDM-DA \citep{tian2022vdm} constructs a proxy source domain by randomly sampling features from the following GMM, 
\setlength\abovedisplayskip{0.0cm}
\setlength\belowdisplayskip{0.2cm}
\begin{equation}
    p_v(z) = \sum\nolimits_{c=1}^{C} \pi_c\; \mathcal{N}(z|\mu_c,\Sigma_c),
\end{equation}
where $z$ denotes the virtual domain feature, and $p_v(z)$ is the distribution of the virtual domain in the feature space.
For each Gaussian component, $\pi_c \ge 0$ represents the mixing coefficient satisfying $\sum_c \pi_c=1$, and $\mu_c,\Sigma_c$ represent the mean and the covariance matrix, respectively.
Specifically, $\mu_c$ is approximated by the $L_2$-normalized class prototype \citep{chen2018closer} that corresponds to the $c$-th row of weights in the source classifier, and a class-agnostic covariance matrix is heuristically determined by pairwise distances among different class prototypes. 
To incorporate relevant knowledge from the target domain, SFDA-DE \citep{ding2022source} further selects confident pseudo-labeled target samples and re-estimates the mean and covariance over these source-like samples as an alternative.
In contrast, CPGA \citep{qiu2021source} trains a prototype generator from conditional noises to generate multiple avatar feature prototypes for each class, encouraging that class prototypes are intra-class compact and inter-class separated.

\method{Virtual domain alignment}
Once the source distribution is estimated, it is essential to seek virtual domain alignment between the proxy source domain and the target domain for knowledge transfer.
We review a variety of virtual domain alignment techniques as follows.
Firstly, SHOT++ \citep{liang2021source} and ProxyMix \citep{ding2022proxymix} follow a classic semi-supervised approach, MixMatch \citep{berthelot2019mixmatch}, to bridge the domain gap.
Secondly, SDDA \citep{kurmi2021domain} adopts the widely-used domain adversarial alignment technique \citep{ganin2015unsupervised} that is formally written as:
\setlength\abovedisplayskip{0.2cm}
\setlength\belowdisplayskip{0.0cm}
\begin{equation}
    \min_{\theta_H}\max_{\theta_D}\ \mathbb{E}_{x_t \in \mathcal{X}_p}[\log D(H(x_t))] + \mathbb{E}_{x_t \in \mathcal{X}_t} [\log(1 - D(H(x_t)))],
\end{equation}
where $H$ and $D$ respectively represent the feature encoder and the binary domain discriminator, and $\mathcal{X}_p$ denotes the proxy source domain.
Due to its simplicity, the domain adversarial training strategy has also been utilized in the following works \citep{liu2021source,ye2021source,tian2022vdm}.
Besides, a certain number of following methods \citep{nayak2021mining,yan2021source,stan2021unsupervised} further employ advanced domain adversarial training strategies to achieve better adaptation.
Thirdly, BAIT \citep{yang2021casting} leverages the maximum classifier discrepancy \citep{saito2018maximum} between two classifiers' outputs in an adversarial manner to achieve feature alignment, which has been followed by \citep{tian2023source,chu2022denoised}.
Fourthly, some TTDA methods \citep{ding2022source,zhang2022divide,liu2022self} explore the maximum mean discrepancy (MMD) \citep{gretton2012kernel} and propose various conditional variants to reduce the difference of features across domains.
In addition, features from different domains could be also aligned through contrastive learning between source prototypes and target samples \citep{qiu2021source,zhang2022divide}.
To model the instance-level alignment, MTRAN \citep{huang2022relative} reduces the difference between features from the target data and its corresponding variant in the virtual source domain.

\subsubsection{Self-supervised Learning}
Self-supervised learning is a learning paradigm tailored to learn feature representation from unlabeled data based on pretext tasks \citep{gidaris2018unsupervised,caron2018deep,caron2020unsupervised,he2020momentum,chen2020simple}.
As mentioned earlier, the centroid-based pseudo labels are similar to the learning manner of DeepCluster \citep{caron2018deep}.
Inspired by rotation prediction \citep{gidaris2018unsupervised}, SHOT++ \citep{liang2021source} further comes up with a relative rotation prediction task and introduces an additional 4-way classification head during adaptation.
Besides, OnTA \citep{wang2021ontarget} and CluP \citep{conti2022cluster} exploit the self-supervised learning frameworks \citep{he2020momentum,caron2020unsupervised} for learning discriminative features as initialization, respectively.
TTT++ \citep{liu2021ttt++} learns an extra self-supervised branch using contrastive learning \citep{chen2020simple} in the source model, which facilitates the adaptation in the target domain with the same objective.
\tim{FedICON \citep{tan2023heterogeneity} leverages unsupervised contrastive learning to guide the model to smoothly generalize to test data under intra-client heterogeneity.}
Recently, StickerDA \citep{kundu2022concurrent} designs three self-supervised objectives such as sticker location, and optimizes the sticker intervention-based pretext task with the auxiliary classification head in both the source training and target adaptation phases.

\method{Remarks}
In addition, some remaining TTDA methods have not been covered in the previous discussions.
PCT \citep{tanwisuth2021prototype} \tim{and POUF \citep{tanwisuth2023pouf} treat} the weights in the classifier layer as source prototypes, and develop an optimal transport-based feature alignment strategy between target features and source prototypes.
Besides, target prototypes could also be considered representative labeled data, and such a prototypical augmentation helps correct the classifier with pseudo-labeling \citep{xiong2022source}. 
LA-VAE \citep{yang2021model} exploits the variational auto-encoder to achieve latent feature alignment.
In addition, the meta-learning mechanism is adopted in a few studies \citep{wang2021self,bohdal2022feed} for the TTDA problem. 
A recent work \citep{naik2023machine} even generates common sense rules and adapts models to the target domain to reduce rule violations.

\subsection{Learning Scenarios of TTDA Algorithms}
\label{sec:ttda-task}
\method{Closed-set \vs Open-set}
Most existing TTDA methods focus on a closed-set scenario, \ie, $\mathcal{C}_s = \mathcal{C}_t$, and some TTDA algorithms \citep{liang2020we,huang2021model} are also validated in a relaxed partial-set setting \citep{liang2020balanced}, \ie, $\mathcal{C}_t \subset \mathcal{C}_s$.
However, several TTDA works \citep{liang2020we,kundu2020towards,feng2021open} consider the open-set learning scenario where the target label space $\mathcal{C}_t$ subsumes the source label space $\mathcal{C}_s$.
To allow more flexibility, open-partial-set domain adaptation \citep{you2019universal} ($\mathcal{C}_s \setminus \mathcal{C}_t \ne \emptyset, \mathcal{C}_t \setminus \mathcal{C}_s \ne \emptyset$,) is studied in TTDA methods \citep{kundu2020universal,deng2021universal,yang2023one}.
Moreover, several recent studies \citep{liang2021umad,qu2023upcycling} even develop a unified framework for both open-set and open-partial-set scenarios.

\method{Single-source \vs Multi-source}
To fully transfer knowledge from multiple source models, prior TTDA methods \citep{liang2020we,liang2021source,kundu2022balancing} extend the single-source TTDA algorithms by combining these adapted models together in the target domain.
Besides, a couple of works \citep{ahmed2021unsupervised,dong2021confident} are elaborately designed for adaptation with multiple source models.
While each source domain typically shares the same label space with the target domain, UnMSMA-MiFL \citep{li2022union} considers a union-set multi-source scenario where the union set of the source label spaces is the same as the target label space.

\method{Single-target \vs Multi-target}
Several TTDA methods \citep{ahmed2022cleaning,kumar2023conmix} also validate the effectiveness of their proposed methods for multi-target domain adaptations where multiple unlabeled target domains exist at the same time.
It is worth noting that each target domain may come in a streaming manner, thus the model is successively adapted to different target domains \citep{rostami2021lifelong,panagiotakopoulos2022online}.

\method{Unsupervised \vs Semi-supervised}
Some TTDA methods \citep{wang2021learning,ma2022context} adapt the source model to the target domain with only a few labeled target samples and adequate unlabeled target samples.
In these semi-supervised learning scenarios, the standard classification loss over the labeled data could be readily incorporated to enhance the adaptation performance \citep{liang2021source,wang2021learning}. 

\method{White-box \vs Black-box}
Sharing a model with all the parameters may not be flexible for adjustment if the model turns out to have harmful applications \footnote{\url{https://openai.com/blog/openai-api/}}. 
In this case, the source model is accessible as a black-box module through the cloud application programming interface (API).
At an early time, IterLNL \citep{zhang2021unsupervised} treats this black-box TTDA problem as learning with noisy labels, and DINE \citep{liang2022dine} develops several structural regularizations within the knowledge distillation framework.
These approaches inspire many recent black-box TTDA works \citep{sun2022prior,peng2022toward,liu2022self,yang2023divide}.
Beyond the deep learning framework, several shallow studies \citep{chidlovskii2016domain,clinchant2016transductive} focus on the black-box TTDA problem with the target features and their predictions available.

\method{Data \vs Label shifts}
Different from TTDA methods that narrowly focus on adaptation under data distribution change $p_\mathcal{S}(x) \not = p_\mathcal{T}(x)$, another family of TTA methods studies label distribution change, $p_\mathcal{S}(y) \not = p_\mathcal{T}(y)$. 
For instance, Saerens \etal \citep{saerens2002adjusting} propose a well-known prior adaptation framework that adapts an off-the-shelf classifier to a new label distribution with unlabeled data at test time, followed by \citep{lipton2018detecting,alexandari2020maximum}.
We refer interested readers to relevant literature \citep{sipka2022hitchhiker}.
A few methods such as ISFDA \citep{li2021imbalanced} and APA \citep{sun2023domain} pay attention to the class-imbalanced TTDA scenario where both data and label shifts are present.

\method{Active TTDA}
To improve the limited performance gains, MHPL \citep{wang2022active} introduces a new setting, active TTDA, where a few target data can be selected to be labeled by human annotators.
This active setting is also studied by other methods \citep{li2022source_mm,kothandaraman2023salad}, and the key point lies in how to select valuable target samples for labeling.

\method{Miscellaneous TTDA scenarios}
In addition, researchers also focus on other aspects of TTDA, \eg, the robustness against adversarial attacks \citep{agarwal2022unsupervised}, the forgetting of source knowledge \citep{yang2021generalized,liu2023twofer}, and the vulnerability to membership inference attack \citep{an2022privacy} and image-agnostic attacks (\eg, blended backdoor attack) \citep{sheng2023adaptguard}.

\section{Test-Time Batch Adaptation}
\label{sec:ttba}
During the testing phase, it is possible that there may exist a single instance or instances from different distributions.
This situation necessitates the development of techniques that can adapt off-the-shelf models to individual instances.
To be concise, we refer to this learning scheme as \emph{test-time instance adaptation} (\aka, standard test-time training \citep{sun2020test} and one-sample generalization \citep{dinnocente2019learning}), which can be viewed as a special case of test-time domain adaptation ($n_t=1$). 

\subsection{Problem Definition}
\begin{definition}[Test-Time Instance Adaptation, TTIA]
Given a classifier $f_\mathcal{S}$ learned on the source domain $\mathcal{D}_\mathcal{S}$, and an unlabeled target instance $x_t \in \mathcal{D}_\mathcal{T}$ under distribution shift, \emph{test-time instance adaptation} aims to leverage the labeled knowledge implied in $f_\mathcal{S}$ to infer the label of $x_t$ adaptively.
\end{definition}

To the best of our knowledge, the concept \emph{test-time adaptation} is first introduced by Wegmann \etal 
\citep{wegmann1998dragon} in 1998, where the speaker-independent acoustic model is adapted to a new speaker with unlabeled data at test time.
However, this differs from the definition of \emph{test-time instance adaptation} mentioned earlier, as it involves using a few instances instead of a single instance for personalized adaptation.
This scenario is frequently encountered in real-world applications, such as in single-image models that are tested on real-time video data \citep{brahmbhatt2018geometry,azimi2022self}.
To avoid ambiguity, we further introduce a generalized learning scheme, \emph{test-time batch adaptation}, and give its definition as follows.

\begin{definition}[Test-Time Batch Adaptation, TTBA]
Given a classifier $f_\mathcal{S}$ learned on the source domain $\mathcal{D}_\mathcal{S}$, and a mini-batch of unlabeled target instances $\{x_t^1, x_t^2, \cdots, x_t^B\} (B\geq 1)$ from $\mathcal{D}_\mathcal{T}$ under distribution shift, \emph{test-time batch adaptation} aims to leverage the labeled knowledge implied in $f_\mathcal{S}$ to infer the label of each instance at the same time. 
\end{definition}

It is important to acknowledge that the inference of each instance is not independent, but rather influenced by the other instances in the mini-batch.
Test-Time Batch Adaptation (TTBA) can be considered a form of TTDA \citep{liang2020we} when the batch size $B$ is sufficiently large. 
Conversely, when the batch size $B$ is equal to 1, TTBA degrades to TTIA \citep{sun2020test}.
Typically, these schemes assume no access to the source data or the ground-truth labels of data on the target distribution.
In the following, we provide a taxonomy of TTBA (including TTIA) algorithms, as well as the learning scenarios.

\setlength{\tabcolsep}{6.0pt}
\begin{table}[!t]
\caption{A taxonomy on TTBA methods with representative strategies.}
\resizebox{0.45\textwidth}{!}{
    \begin{tabular}{ll}
        \toprule
        \textbf{Families} & \textbf{Representative Strategies}\\
        \midrule
        \textbf{BN calibration} & PredBN \citep{nado2020evaluating,schneider2020improving},  InstCal \citep{zou2022learning}\\
        \textbf{model optimization} & TTT \citep{sun2020test}, GeOS \citep{dinnocente2019learning}, MEMO \citep{zhang2022memo} \\
        \textbf{meta-learning} & MLSR \citep{park2020fast}, Full-OSHOT \citep{borlino2022self} \\
        \textbf{input adaptation} & TPT \citep{shu2022test}, TTA-DAE \citep{karani2021test} \\
        \textbf{dynamic inference} & LAME \citep{boudiaf2022parameter}, EMEA \citep{wang2021efficient} \\
        \bottomrule
\end{tabular}}
\end{table}

\subsection{Taxonomy on TTBA Algorithms}
\subsubsection{Batch Normalization Calibration}
Normalization layers (\eg, batch normalization \citep{ioffe2015batch} and layer normalization \citep{ba2016layer}) are considered essential components of modern neural networks.
For example, a batch normalization (BN) layer calculates the mean and variance for each activation over the training data $\mathcal{X}_\mathcal{S}$, and normalizes each incoming sample $x_s$ as follows,
\setlength\abovedisplayskip{0.0cm}
\setlength\belowdisplayskip{0.2cm}
\begin{equation}
    \hat{x}_s = \gamma \cdot \frac{x_s - \mathbb{E}[X_\mathcal{S}]}{\sqrt{\mathbb{V}[X_\mathcal{S}] + \epsilon}} + \beta,
\end{equation}
where $\gamma$ and $\beta$ denote the scale and shift parameters (\aka, the learnable affine transformation parameters), and $\epsilon$ is a small constant introduced for numerical stability. 
The BN statistics (\ie, the mean $\mu_s = \mathbb{E}[\mathcal{X}_\mathcal{S}]$ and variance $\sigma^2_s = \mathbb{V}[\mathcal{X}_\mathcal{S}]$) are typically approximated using EMA over batch-level estimates $\{\hat{\mu}_k,\hat{\sigma}^2_k\}$, 
\setlength\abovedisplayskip{0.2cm}
\setlength\belowdisplayskip{0.2cm}
\begin{equation}
    \mu_{s} \gets (1-\rho) \cdot\mu_{s} + \rho \cdot\hat{\mu}_k,\; \sigma^2_{s} \gets (1-\rho) \cdot\sigma^2_{s} + \rho \cdot\hat{\sigma}^2_k,
\end{equation}
where $\rho$ is the momentum, $k$ denotes the training step, and the statistics over the $k$-th mini-batch $\{x_i\}_{i=1}^{B_s}$ are
\setlength\abovedisplayskip{0.0cm}
\setlength\belowdisplayskip{0.2cm}
\begin{equation}
\hat{\mu}_{k} = \frac{1}{B_s} \sum\nolimits_i x_i,\; \hat{\sigma}^2_{k}=\frac{1}{B_s} \sum\nolimits_i (x_i - \mu_{k})^2,
\label{eq:bn}
\end{equation}
where $B_s$ denotes the batch size at training time.
During inference, the BN statistics estimated at training time are frozen for each test sample.
AdaBN \citep{li2017revisiting}, a seminal work in the DA literature, suggests that the statistics in the BN layers represent domain-specific knowledge.
To bridge the domain gap, AdaBN replaces the training BN statistics with new statistics estimated over the entire target domain. 
PredBN \citep{nado2020evaluating}, a pioneering TTBA method, substitutes the training BN statistics with those estimated per test batch.

PredBN+ \citep{schneider2020improving} adopts the running averaging strategy for BN statistics during training and suggests mixing the BN statistics per batch with the training statistics $\{\mu_s,\sigma^2_s\}$ as, 
\setlength\abovedisplayskip{0.2cm}
\setlength\belowdisplayskip{0.2cm}
\begin{equation}
    \bar{\mu}_{t} = (1-\rho_t) \cdot\mu_{s} + \rho_t \cdot\hat{\mu}_t,\; \bar{\sigma}^2_{t} = (1-\rho_t) \cdot\sigma^2_{s} + \rho_t \cdot\hat{\sigma}^2_t,
\end{equation}
where the test statistics $\{\hat{\mu}_t,\hat{\sigma}^2_t\}$ are estimated via Eq.~(\ref{eq:bn}), and the hyper-parameter $\rho_t$ controls the trade-off between training and estimated test statistics.
Moreover, TTN \citep{lim2023ttn} presents an alternative solution that calibrates the estimation of the variance as follows,
\setlength\abovedisplayskip{0.2cm}
\setlength\belowdisplayskip{0.2cm}
\begin{equation}
   \bar{\sigma}^2_{t} = (1-\rho_t) \cdot\sigma^2_{s} + \rho_t \cdot\hat{\sigma}^2_t + \rho_t (1-\rho_t) (\hat{\mu}_t - \mu_s)^2.
\end{equation}
Instead of using the same value for different BN layers, TTN optimizes the interpolating weight $\rho_t$ during the post-training phase using labeled source data.
\tim{Alternatively, DN \citep{zhou2023test} proposes subtracting the mean of embeddings within each mini-batch before inference.}

Typically, methods that rectify BN statistics may suffer from limitations when the batch size $B$ is small, particularly when $B=1$.
SaN \citep{bahmani2022adaptive} directly attempts to mix instance normalization (IN) \citep{ulyanov2016instance} statistics estimated per instance with the training BN statistics.
Instead of manually specifying a fixed value at test time, InstCal \citep{zou2022learning} introduces an additional module during training to learn the interpolating weight between IN and BN statistics, allowing the network to dynamically adjust the importance of training statistics for each test instance.
By contrast, AugBN \citep{khurana2021sita} expands a single instance to a batch of instances using random augmentation, then estimates the BN statistics using the weighted average over these augmented instances.

\subsubsection{Model Optimization}
Another family of TTBA methods involves adjusting the parameters of a pre-trained model for each unlabeled test instance (batch). 
These methods are generally divided into two main categories: (1) training with auxiliary tasks \citep{dinnocente2019learning,sun2020test,dinnocente2020one}, which introduces an additional self-supervised learning task in the primary task during both training and test phases, and (2) fine-tuning with unsupervised objectives \citep{wang2019dynamic,zhang2022memo,reddy2022master}, which elaborately designs a task-specific objective for updating the pre-trained model.

\method{Training with auxiliary tasks}
Motivated by prior works \citep{carlucci2019domain,sun2019unsupervised} in which incorporating self-supervision with supervised learning in a unified multi-task framework enhances adaptation and generalization, TTT \citep{sun2020test} and OSHOT \citep{dinnocente2020one} are two pioneering works that leverage the same self-supervised learning (SSL) task at both training and test phases, to implicitly align features from the training domain and the test instance.
Specifically, they adopt a common multi-task architecture, comprising the primary classification head $h_c(\cdot;\theta_c)$, the SSL head $h_s(\cdot;\theta_s)$, and the shared feature encoder $f_e(\cdot;\theta_e)$.
The following joint objective of TTT or OSHOT is optimized at the training stage,
\setlength\abovedisplayskip{0.1cm}
\setlength\belowdisplayskip{0.1cm}
\begin{equation}
   \theta_e^{*}, \theta_c^{*}, \theta_s^{*} = \mathop{\arg\min}_{\theta_e, \theta_c, \theta_s} \sum_{i=1}^{n_s} \mathcal{L}_{pri}(x_i,y_i;\theta_e,\theta_c) + \mathcal{L}_{ssl}(x_i;\theta_e,\theta_s),
\label{eq:ttt}
\end{equation}
where $\mathcal{L}_{pri}$ denotes the primary objective (\eg, cross-entropy for classification tasks), and $\mathcal{L}_{ssl}$ denotes the auxiliary SSL objective (\eg, rotation prediction \citep{gidaris2018unsupervised} and solving jigsaw puzzles \citep{carlucci2019domain}).
For each test instance $x_t$, TTT \citep{sun2020test} first adjusts the feature encoder $f_e(\cdot;\theta_e)$ by optimizing the SSL objective,
\setlength\abovedisplayskip{0.2cm}
\setlength\belowdisplayskip{0.2cm}
\begin{equation}
   \theta_e(x_t) = \mathop{\arg\min}_{\theta_e} \mathcal{L}_{ssl}(x_t;\theta_s^{*},\theta_e),
\end{equation}
then obtains the prediction with the adjusted model as $\hat{y} = h_c(f_e(x;\theta_e(x_t));\tim{\theta_c^{*}})$.
By contrast, OSHOT \citep{dinnocente2020one} modifies the parameters of both the feature encoder and the SSL head according to the SSL objective at test time.
Generally, many follow-up methods adopt the same auxiliary training strategy by developing various self-supervisions for different applications \citep{zhang2020inference,hansen2021self,gandelsman2022test}.
Among them, TTT-MAE \citep{gandelsman2022test} is a recent extension of TTT that utilizes the transformer backbone and replaces the self-supervision with masked autoencoders \citep{he2022masked}.

To increase the dependency between the primary task and the auxiliary task, GeOS \citep{dinnocente2019learning} further adds the features of the SSL head to the primary head.
SR-TTT \citep{lyu2022learning} does not follow the Y-shaped architecture but instead utilizes an explicit connection between the primary task and the auxiliary task.
Specifically, SR-TTT takes the output of the primary task as the input of the auxiliary task.
TTCP \citep{sarkar2022leveraging} follows the same pipeline as TTT, but it leverages a test-time prediction ensemble strategy by identifying augmented samples that the SSL head could correctly classify.

\method{Training-agnostic fine-tuning}
To avoid modifying training with auxiliary tasks in the source domain, the other methods focus on developing unsupervised objectives solely for optimizing the model at test time. 
DIEM \citep{wang2019dynamic} proposes a selective entropy minimization objective for pixel-level semantic segmentation, while MALL \citep{reddy2022master} enforces edge consistency prior through a weighted normalized cut loss.
Besides, MEMO \citep{zhang2022memo} optimizes the entropy of the averaged prediction over multiple random augmentations of the input sample.
\tim{PromptAlign \citep{samadh2023align} additionally handles the train-test distribution shift by matching the mean and variances of the test sample and the source dataset statistics.}
TTAS \citep{bateson2022test} further develops a class-weighted entropy objective, while SUTA \citep{lin2022listen} additionally incorporates minimum class confusion to reduce the uncertainty.
\tim{A recent work \citep{zhao2024test} develops a reinforcement learning approach that updates the model parameters via policy gradient to maximize the expected reward.}

Self-supervised consistency regularization under various input variations is also favorable in customizing the pre-trained model for each test input \citep{liu2022single,jin2023empowering}.
In particular, SCIO \citep{kan2022self} develops a self-constrained optimization method to learn the coherent spatial structure.
While adapting image models to a video input \citep{brahmbhatt2018geometry,li2020online}, ensuring temporal consistency between adjacent frames is a crucial aspect of the unsupervised learning objective.
Many other methods directly update the model with the unlabeled objectives tailored to specific tasks, \eg, image matching \citep{hong2021deep}, image denoising \citep{mohan2021adaptive}, generative modeling \citep{bau2019semantic}, and style transfer \citep{kim2022controllable}.
In addition, the model could be adapted to each instance by utilizing the generated data at test time.
As an illustration, TTL-EQA \citep{banerjee2021self} generates numerous synthetic question-answer pairs and subsequently leverages them to infer answers in the given context.
ZSSR \citep{shocher2018zero} trains a super-resolution network using solely down-sampled examples extracted from the test image itself.

\subsubsection{Meta-Learning}
MAML \citep{finn2017model}, a notable example of meta-learning \citep{hospedales2021meta}, learns a meta-model that can be quickly adapted to perform well on a new task using a small number of samples and gradient steps.
Such a learning paradigm is typically well-suited for test-time adaptation, where we can update the meta-model using an unlabeled objective over a few test data.
There exist two distinct categories: backward propagation \citep{park2020fast,borlino2022self}, and forward propagation \citep{dubey2021adaptive,kim2022variational}. 
The latter category does not alter the trained model but includes the instance-specific information in the dynamical neural network.

\method{Backward propagation} Inspired by the pioneering work \citep{shocher2018zero}, MLSR \citep{park2020fast} develops a meta-learning method based on MAML for single-image super-resolution.
Concretely, the meta-objective \wrt the network parameter $\theta$ is shown as,
\setlength\abovedisplayskip{0.2cm}
\setlength\belowdisplayskip{0.2cm}
\begin{equation}
  \min_{\theta} \sum\nolimits_i \mathcal{L}(\text{LR}_i,\text{HR}_i; \theta - \alpha {\nabla_\theta \mathcal{L}(\text{LR}_i\downarrow,\text{LR}_i;\theta})),
  \label{eq:meta-train}
\end{equation}
where $\mathcal{L}(A, B; \theta)=\|f_\theta(A)-B\|_2^2$ is the loss function, $\alpha$ is the learning rate of gradient descent, and $\text{LR}_i\downarrow$ denotes the down-scaled version of the low-resolution input in the paired trained data $(\text{LR}_i,\text{HR}_i)$. 
At inference time, MLSR first adapts the meta-learned network to the low-resolution test image and its down-sized image (LR$\downarrow$) using the parameter $\theta^*$ learned in Eq.~(\ref{eq:meta-train}) as initialization,
\begin{equation}
  \theta_t \gets \theta^* - \alpha {\nabla_\theta \mathcal{L}(\text{LR}\downarrow,\text{LR};\theta^*)},
  \label{eq:meta-test}
\end{equation}
then generates the high-resolution (HR) image as $f_{\theta_t}(\text{LR})$.
Such a meta-learning mechanism based on self-supervised learning has been utilized by follow-up methods \citep{chi2021test,liu2022towards,min2023meta}.
Among them, MetaVFI \citep{choi2021test} further introduces self-supervised cycle consistency for video frame interpolation.

As an alternative, Full-OSHOT \citep{borlino2022self} proposes a meta-auxiliary learning approach that optimizes the shared encoder with an inner auxiliary task, providing a better initialization for the subsequent primary task:
\setlength\abovedisplayskip{0.0cm}
\setlength\belowdisplayskip{0.2cm}
\begin{equation}
   \min_{\theta_e, \theta_c} \sum\nolimits_i \mathcal{L}_{pri}(x_i,y_i;\theta_e - \alpha {\nabla_{\theta_e} \mathcal{L}_{ssl}(x_i;\theta_e,\theta_s)}, \theta_c),
\end{equation}
and the definitions of variables are the same as OSHOT \citep{dinnocente2020one} in Eq.~(\ref{eq:ttt}).
After the meta-training phase, the parameters $(\theta_e, \theta_s)$ are updated for each test sample according to the auxiliary self-supervised objective.
This learning paradigm is also known as meta-tailoring \citep{alet2021tailoring}, where $\mathcal{L}_{ssl}$ in the inner loop affects the optimization of $\mathcal{L}_{pri}$ in the outer loop.
Subsequent methods exploit various self-supervisions in the inner loop, including contrastive learning \citep{alet2021tailoring} and reconstruction \citep{sain2022sketch3t,liu2023meta}.

\method{Forward propagation} Apart from the shared encoder $f_e(\theta_e)$ above, several other meta-learning methods exploit the normalization \tim{statistics \citep{zhang2021adaptive,bao2023adaptive}} or domain prototypes \citep{dubey2021adaptive,kim2022variational} from the inner loop, allowing backward-free adaptation at inference time. 
Besides, some works incorporate extra meta-adjusters \citep{sun2022dynamic} or learnable prompts \citep{ben2022pada}, by taking the instance embedding as input, to dynamically generate a small subset of parameters in the network, which are optimized at the training phase.
DSON \citep{seo2020learning} proposes to fuse IN with BN statistics by linearly interpolating the means and variances, incorporating the instance-specific information in the trained model.
Following another popular meta-learning framework \citep{li2019episodic}, SSGen \citep{xiao2022learning} suggests episodically dividing the training data into meta-train and meta-test to learn the meta-model, which is subsequently applied to the entire training data for final test-time inference. 
It is also employed by \citep{xu2022mimic,segu2023batch} where multiple source domains are involved during training.

\subsubsection{Input Adaptation}
In contrast to model-level optimization, which updates pre-trained models for input data, another line of TTBA methods focuses on changing input data for pre-trained models \citep{karani2021test,zhao2022test,gao2023back}.
For example, \tim{TPT \citep{shu2022test} freezes the pre-trained multimodal model and only learns the extra text prompt based on the marginal entropy of each instance. Another approach,} CVP \citep{tsai2023self}, optimizes the convolutional visual prompts in the input under the guidance of a self-supervised contrastive learning objective.

TTA-AE \citep{he2021autoencoder} additionally learns a set of auto-encoders in each layer of the trained model at training time.
It is posited that unseen inputs have larger reconstruction errors than seen inputs, thus a set of domain adaptors is introduced at test time to minimize the reconstruction loss.
Similarly, TTA-DAE \citep{karani2021test} only learns an image-to-image translator (\aka, input adaptor) for each input so that the frozen training-time denoising auto-encoder could well reconstruct the network output.
TTO-AE \citep{li2022self} follows the Y-shaped architecture of TTT and optimizes both the shared encoder and the additional input adaptor to minimize reconstruction errors in both heads.
Instead of auxiliary auto-encoders, AdvTTT \citep{valvano2022reusing} leverages a discriminator that is adversarially trained to distinguish real from predicted network outputs, so that the prediction output for each adapted test input satisfies the adversarial output prior.

OST \citep{termohlen2021continual} proposes mapping the target input onto the source data manifold using Fourier style transfer \citep{yang2020fda}, serving as a pre-processor to the primary network.
By contrast, TAF-Cal \citep{zhao2022test} further utilizes the average amplitude feature over the training data to perform Fourier style calibration \citep{yang2020fda} at both training and test phases, bridging the gap between training and test data.
It is noteworthy that imposing a data manifold constraint \citep{pandey2021generalization,sarkar2022leveraging,gao2023back,xiao2023energy} can aid in achieving better alignment between the test data and unseen training data. 
Specifically, ITTP \citep{pandey2021generalization} trains a generative model over source features with target features projected onto points in the source feature manifold for final inference.
DDA \citep{gao2023back} exploits the generative diffusion model for target data, while ESA \citep{xiao2023energy} updates the target feature by energy minimization through Langevin dynamics.

In addition to achieving improved recognition results against domain shifts, a certain number of TTBA methods also explore input adaptation for the purpose of test-time adversarial defense \citep{shi2021online,yoon2021adversarial,mao2021adversarial,alfarra2022combating}.
Among them, Anti-Adv \citep{alfarra2022combating} perturbs the test input to maximize the classifier's prediction confidence.
Besides, SOAP \citep{shi2021online} leverages self-supervisions like rotation prediction at both training and test phases and purifies adversarial test examples based on self-supervision only.
SSRA \citep{mao2021adversarial} only exploits the self-supervised consistency under different augmentations at test time to remove adversarial noises in the attacked data.

\subsubsection{Dynamic Inference}
LAME \citep{boudiaf2022parameter} utilizes neighbor consistency to enforce consistent assignments on neighboring points in the feature space, without modifying the pre-trained model.
Upon multiple pre-trained models learned from the source data, a few works \citep{wang2021efficient,zhang2023domain} learn the weights for each model, without making any changes to the models themselves. 
For example, EMEA \citep{wang2021efficient} employs entropy minimization to update the ensemble coefficients before each model.
GPR \citep{jain2011online} is one of the early works that only adjusts the network predictions instead of the pre-trained model.
In particular, it bootstraps the more difficult faces in an image from the more easily detected faces and adopts Gaussian process regression to encourage smooth predictions for similar patches.

\subsection{Learning Scenarios of TTBA Algorithms}
\method{Instance \vs Batch}
As defined above, test-time adaptation could be divided into two cases: instance adaptation \citep{sun2020test,zhang2022memo} and batch adaptation \citep{schneider2020improving,brahmbhatt2018geometry}, according to whether a single instance or a batch of instances exist at test time. 

\method{Single \vs Multiple}
In contrast to vanilla test-time adaptation that utilizes the pre-trained model from one single source domain, some works (\eg,  \citep{dinnocente2019learning,pandey2021generalization,wang2021efficient,xiao2022learning,zhao2022test,xiao2023energy,zhang2023domain}) are interested in domain generalization problems where multiple source domains exist.

\method{White-box \vs Black-box}
A majority of TTBA methods focus on adapting white-box models to test instances, while some other works (\eg, \citep{jain2011online,chen2019self,zhang2023domain}) do not have access to the parameters of the pre-trained model (black-box) and instead adjust the predictions according to generic structural constraints.

\method{Customized \vs On-the-fly}
Most existing TTA methods require training one or more customized models in the source domain, \eg, TTT \citep{sun2020test} employs a Y-shaped architecture with an auxiliary head.
However, it may be not allowed to train the source model in a customized manner for some real-world applications.
Other works \citep{zhang2022memo,alfarra2022combating} do not rely on customized training in the source domain but develop flexible techniques for adaptation with on-the-fly models. 

\section{Online Test-Time Adaptation}
\label{sec:otta}
Previously, we have considered various test-time adaptation scenarios where pre-trained source models are adapted to a domain \citep{liang2020we,li2020model}, a mini-batch \citep{schneider2020improving,zhang2021adaptive}, or even a single instance \citep{sun2020test,zhang2022memo} at test time.
However, offline test-time adaptation typically requires a certain number of samples to form a mini-batch or a domain, which may be infeasible for streaming data scenarios where data arrives continuously and in a sequential manner.
To reuse past knowledge like online learning, TTT \citep{sun2020test} employs an online variant that does not optimize the model episodically for each input but instead retains the optimized model for the last input.

\subsection{Problem Definition}
\begin{definition}[Online Test-Time Adaptation, OTTA]
Given a well-trained classifier $f_\mathcal{S}$ on the source domain $\mathcal{D}_\mathcal{S}$ and a sequence of unlabeled mini-batches $\{\mathcal{B}_1, \mathcal{B}_2, \cdots\}$, \emph{online test-time adaptation} aims to leverage the labeled knowledge implied in $f_\mathcal{S}$ to infer labels of samples in $\mathcal{B}_i$ under distribution shift, in an online manner. 
In other words, the knowledge learned in previously seen mini-batches could be accumulated for adaptation to the current mini-batch.
\end{definition}

The above definition corresponds to the problem addressed in Tent \citep{wang2021tent}, where multiple mini-batches are sampled from a new data distribution that is distinct from the source data distribution.
Besides, it also encompasses the online test-time instance adaptation problem, as introduced in TTT-Online \citep{sun2020test} when the batch size equals 1.
However, samples at test time may come from a variety of different distributions, leading to new challenges such as error accumulation and catastrophic forgetting.
To address this issue, CoTTA \citep{wang2022continual} and EATA \citep{niu2022efficient} investigate the continual test-time adaptation problem that adapts the pre-trained source model to the continually changing test data. 
Such a non-stationary adaptation problem could be also viewed as a special case of the definition above, when each mini-batch may come from a different distribution.

\setlength{\tabcolsep}{4.0pt}
\begin{table}[!t]
\caption{A taxonomy on OTTA methods with representative strategies.}
\resizebox{0.47\textwidth}{!}{
    \begin{tabular}{ll}
        \toprule
        \textbf{Families} & \textbf{Representative Strategies}\\
        \midrule
        \textbf{BN calibration} & DUA \citep{mirza2022norm}, DELTA \citep{zhao2023delta}\\
        \textbf{entropy minimization} & Tent \citep{wang2021tent}, SAR \citep{niu2023towards} \\
        \textbf{pseudo-labeling} & T3A \citep{iwasawa2021test}, TAST \citep{jang2023test}\\
        \textbf{consistency regularization} & CFA \citep{kojima2022robustifying}, PETAL \citep{brahma2023probabilistic} \\
        \textbf{anti-forgetting regularization} & CoTTA \citep{wang2022continual}, EATA \citep{niu2022efficient} \\
        \bottomrule
\end{tabular}}
\end{table}

\subsection{Taxonomy on OTTA Algorithms}
\subsubsection{Batch Normalization Calibration} 
As noted in the previous section, normalization layers such as batch normalization (BN) \citep{ioffe2015batch} are commonly employed in modern neural networks. 
Typically, BN layers can encode domain-specific knowledge into normalization statistics \citep{li2017revisiting}.
A recent work \citep{niu2023towards} further investigates the effects of different normalization layers under the test-time adaptation setting. In the following, we mainly focus on the BN layer due to its widespread usage in existing methods.

Tent \citep{wang2021tent} and RNCR \citep{hu2021fully} propose replacing the fixed BN statistics (\ie, mean and variance $\{\mu_s, \sigma^2_s\}$) in the pre-trained model with the estimated ones $\{\hat{\mu}_t, \hat{\sigma}^2_t\}$ from the $t$-th test batch.
CD-TTA \citep{song2022cdtta} develops a switchable mechanism that selects the most similar one from multiple BN branches in the pre-trained model using the Bhattacharya distance.
Besides, Core \citep{you2021test} calibrates the BN statistics by interpolating between the fixed source statistics and the estimated ones at test time, namely, $\mu_t = \rho \hat{\mu}_t + (1-\rho) \mu_s, \sigma_t = \rho \hat{\sigma}_t + (1-\rho) \sigma_s$, where $\rho \in [0,1]$ is a momentum hyper-parameter.

Similar to the running average estimation of BN statistics during training, ONDA \citep{mancini2018kitting} proposes initializing the BN statistics $\{\mu_0, \sigma_0^2\}$ as $\{\mu_s, \sigma_s^2\}$ and updating them for the $t$-th test batch,
\setlength\abovedisplayskip{0.2cm}
\setlength\belowdisplayskip{0.2cm}
\begin{equation}
    \begin{aligned}
    \mu_t &= \rho \hat{\mu}_t + (1-\rho) \mu_{t-1},\\
    \sigma_t^2 &= \rho \hat{\sigma}_t^2 + (1-\rho)\frac{n_t}{n_t-1}\sigma_{t-1}^2,
    \end{aligned}
    \label{eq:bn_ema}
\end{equation}
where $n_t$ denotes the number of samples in the batch, and $\rho$ is a momentum hyper-parameter.
Instead of a constant value for $\rho$, MECTA \citep{hong2023mecta} considers a heuristic weight through computing the distance between $\{\mu_{t-1}, \sigma_{t-1}\}$ and $\{\hat{\mu}_t, \hat{\sigma}_t\}$.
\tim{EDTN \citep{wang2024optimization} further introduces a straightforward layer-wise strategy to set the momentum hyper-parameters for different layers.}

To decouple the gradient backpropagation and the selection of BN statistics, GpreBN \citep{yang2022test} and DELTA \citep{zhao2023delta} adopt the following reformulation of batch re-normalization \citep{ioffe2017batch}, 
\setlength\abovedisplayskip{0.2cm}
\setlength\belowdisplayskip{0.2cm}
\begin{equation}
    \hat{x}_t = \gamma \cdot \frac{\frac{x_t-\hat{\mu}_t}{\hat{\sigma}_t} \cdot sg(\hat{\sigma}_t) + sg(\hat{\mu}_t) - \mu}{\sigma}+ \beta,
\end{equation}
where $sg(\cdot)$ denotes the stop-gradient operation, and $\{\gamma,\beta\}$ are the affine parameters in the BN layer.
To obtain stable BN statistics $\{\mu, \sigma^2\}$, these methods utilize the test-time dataset-level running statistics via the moving average like Eq.~(\ref{eq:bn_ema}).

For online adaptation with a single sample, MixNorm \citep{hu2021mixnorm} mixes the estimated IN statistics with the exponential moving average BN statistics at test time.
On the other hand, DUA \citep{mirza2022norm} adopts a decay strategy for the weighting hyper-parameter $\rho$ and forms a small batch from a single image to stabilize the online adaptation process.
To obtain more accurate estimates of test-time statistics, NOTE \citep{gong2022note} maintains a class-balanced memory bank that is utilized to update the BN statistics using an exponential moving average.
Additionally, NOTE proposes a selective mixing strategy that only calibrates the BN statistics for detected out-of-distribution samples.
TN-SIB \citep{zhang2022generalizable} also leverages a memory bank that provides samples with similar styles to the test sample, to accurately estimate BN statistics. 

\subsubsection{Entropy Minimization}
Entropy minimization is a widely used technique to handle unlabeled data. 
A pioneering approach, Tent \citep{wang2021tent}, proposes minimizing the mean entropy over the test batch to update the affine parameters $\{\gamma, \beta\}$ of BN layers in the pre-trained model, followed by various subsequent methods \citep{gong2022note,yang2022test}.
Notably, VMP \citep{jing2022variational} reformulates Tent in a probabilistic framework by introducing perturbations into the model parameters by variational Bayesian inference.
Several other methods \citep{tang2023neuro,yi2023temporal} also focus on minimizing the entropy at test time but utilize different combinations of learnable parameters.
BACS \citep{zhou2021bayesian} incorporates the entropy regularization for unlabeled data in the approximate Bayesian inference algorithm, and samples multiple model parameters to obtain the marginal probability for each sample.
In addition, TTA-PR \citep{sivaprasad2021test} proposes minimizing the average entropy of predictions under different augmentations. 
FEDTHE+ \citep{jiang2023test} employs the same adaptation scheme as MEMO \citep{zhang2022memo} that minimizes the entropy of the average prediction over different augmentations.

To avoid overfitting to non-reliable and redundant test data, EATA \citep{niu2022efficient} develops a sample-efficient entropy minimization strategy that identifies samples with lower entropy values than the pre-defined threshold for model updates, which is also adopted by follow-up methods \citep{song2023ecotta,niu2023towards}.
CD-TTA \citep{song2022cdtta} leverages the similarity between feature statistics of the test sample and source running statistics as sample weights, instead of using discrete weights $\{0,1\}$.
Besides, DELTA \citep{zhao2023delta} derives a class-wise re-weighting approach that associates sample weights with corresponding pseudo labels to mitigate bias towards dominant classes.

There exist many alternatives to entropy minimization for adapting models to unlabeled test samples including class confusion minimization \citep{you2021test}, batch nuclear-norm maximization \citep{hu2021fully}, maximum squares loss \citep{song2022cdtta}, and mutual information maximization \citep{kingetsu2022multi,choi2022improving}.
In addition, MuSLA \citep{kingetsu2022multi} further considers the virtual adversarial training objective that enforces classifier consistency by adding a small perturbation to each sample.
SAR \citep{niu2023towards} encourages the model to lie in a flat area of the entropy loss surface and optimizes the minimax entropy objective below,
\setlength\abovedisplayskip{0.2cm}
\setlength\belowdisplayskip{0.2cm}
\begin{equation}
    \min_{\theta} \max_{\|\Delta_\theta\|_2 \leq \epsilon} \mathcal{H}(x;\theta + \Delta_\theta),
\end{equation}
where $\mathcal{H}(\cdot)$ denotes the entropy function, and $\Delta_\theta$ denotes the weight perturbation in a Euclidean ball with radius $\epsilon$.
Moreover, a few methods \citep{kundu2022uncertainty,yang2023auto} even employ entropy maximization for specific tasks, for example, AUTO \citep{yang2023auto} performs model updating for unknown samples at test time.

\subsubsection{Pseudo-labeling}
Unlike the unidirectional process of entropy minimization, many OTTA methods \citep{belli2022online,kingetsu2022multi,boudiaf2022parameter,song2022cdtta} adopt pseudo labels generated at test time for model updates.
Among them, MM-TTA \citep{shin2022mmtta} proposes a selective fusion strategy to ensemble predictions from multiple modalities.
Besides, DLTTA \citep{yang2022dltta} obtains soft pseudo labels by averaging the predictions of its nearest neighbors in a memory bank, and subsequently optimizes the symmetric KL divergence between the model outputs and these pseudo labels.
TAST \citep{jang2023test} proposes a similar approach that reduces the difference between predictions from a prototype-based classifier and a neighbor-based classifier.
Notably, SLR+IT \citep{mummadi2021test} develops a negative log-likelihood ratio loss instead of the commonly used cross-entropy loss, providing non-vanishing gradients for highly confident predictions.

Conjugate-PL \citep{goyal2022test} presents a way of designing unsupervised objectives for TTA by leveraging the convex conjugate function.
The resulting objective resembles self-training with specific soft labels, referred to as conjugate pseudo labels.
A recent work \citep{wang2023towards} theoretically analyzes the difference between hard and conjugate labels under gradient descent for a binary classification problem.
Motivated by the idea of negative learning \citep{kim2019nlnl}, ECL \citep{han2023rethinking} further considers complementary labels from the least probable categories.
Besides, T3A \citep{iwasawa2021test} proposes merely adjusting the classifier layer by computing class prototypes using online unlabeled data and classifying each unlabeled sample based on its distance to these prototypes.

\subsubsection{Consistency Regularization}
In the classic mean teacher \citep{tarvainen2017mean} framework, the pseudo labels under weak data augmentation obtained by the teacher network are known to be more stable.
Built on this framework, RMT \citep{dobler2023robust} pursues the teacher-student consistency in predictions through a symmetric cross-entropy measure, while OIL \citep{ye2022robust} only exploits highly confident samples during consistency maximization. 
VDP \citep{gan2023decorate} utilizes this framework to update visual domain prompts with the pre-trained model being frozen. 
Moreover, CoTTA \citep{wang2022continual} further employs multiple augmentations to refine the pseudo labels from the teacher network, which is also applied in other methods \citep{brahma2023probabilistic,tomar2023tesla,ma2023swapprompt}.
Inspired by maximum classifier discrepancy \citep{saito2018maximum}, AdaODM \citep{zhang2022adaptive} proposes minimizing the prediction disagreement between two classifiers at test time to update the feature encoder.

Apart from the model variation above, several methods \citep{sivaprasad2021test,das2023transadapt,lumentut20223d,su2022revisiting,chen2023openworld} also enforce the consistency of the corresponding predictions among different augmentations.
In particular, SWR-NSP \citep{choi2022improving} introduces an additional nearest source prototype classifier at test time and minimizes the difference between predictions under two different augmentations.
Besides, many methods \citep{guan2021bilevel,kuznietsov2022towards,kim2022ev,belli2022online,yi2023temporal} leverage the temporal coherence for video data and design a temporal consistency objective at test time.
For example, TeCo \citep{yi2023temporal} encourages adjacent frames to have semantically similar features to increase the robustness against corruption at test time. 

In contrast to constraints in the prediction space, FEDTHE+ \citep{jiang2023test} pursues consistency in the feature space.
Several other OTTA methods \citep{wu2021domainagnostic,dobler2023robust,su2022revisiting}) even pursue consistency between test features and source or target prototypes in the feature space. 
CFA \citep{kojima2022robustifying} further proposes matching multiple central moments to achieve feature alignment. 
Furthermore, ACT-MAD \citep{mirza2022actmad} performs feature alignment by minimizing the discrepancy between the pre-computed training statistics and the estimates of test statistics.
TTAC \citep{su2022revisiting} calculates the online estimates of feature mean and variance at test time instead. 
Besides, CAFA \citep{jung2022cafa} uses the Mahalanobis distance to achieve low intra-class variance and high inter-class variance for test data.

\subsubsection{Anti-forgetting Regularization}
Previous studies \citep{wang2022continual,niu2022efficient} find that the model optimized by TTA methods suffers from severe performance degradation (named forgetting) on original training samples.
To mitigate the forgetting issue, a natural solution is to keep a small subset of training data that is further learned at test time as regularization \citep{belli2022online,dobler2023robust,kuznietsov2022towards}. 
PAD \citep{wu2021domainagnostic} comes up with an alternative approach that keeps the relative relationship of irrelevant auxiliary data unchanged after test-time optimization.
AUTO \citep{yang2023auto} maintains a memory bank to store easily recognized samples for replay and prevents overfitting towards unknown samples at test time.

Another anti-forgetting solution lies in using merely a few parameters for test-time model optimization. 
For example, Tent \citep{wang2021tent} only optimizes the affine parameters in the BN layers for test-time adaptation, and AUTO \citep{yang2023auto} updates the last feature block in the pre-trained model.
SWR-NSP \citep{choi2022improving} divides the entire model parameters into shift-agnostic and shift-biased parameters and updates the former less and the latter more.
Recently, VDP \citep{gan2023decorate} fixes the pre-trained model but only optimizes the input prompts during adaptation.

Besides, CoTTA \citep{wang2022continual} proposes a stochastic restoration technique that randomly restores a small number of parameters to the initial weights in the pre-trained model.
PETAL \citep{brahma2023probabilistic} further selects parameters with smaller gradient norms in the entire model for restoration.
By contrast, EATA \citep{niu2022efficient} introduces an importance-aware Fisher regularizer to prevent excessive changes in model parameters. 
The importance is estimated from test samples with generated pseudo labels.
SAR \citep{niu2023towards} proposes a sharpness-aware and reliable optimization scheme, which removes samples with large gradients and encourages model weights to lie in a flat minimum. 
Further, EcoTTA \citep{song2023ecotta} presents a self-distilled regularization by forcing the output of the test model to be close to that of the pre-trained model.

\method{Remarks}
There are several other solutions for the OTTA problem, \eg, meta-learning \citep{zhang2022generalizable,wu2023learning}, Hebbian learning \citep{tang2023neuro}, and adversarial data augmentation \citep{tomar2023tesla}.
\tim{TDA \citep{karmanov2024efficient} further provides a training-free solution by leveraging a dynamic memory bank that stores pseudo labels and features from previous samples.}

\subsection{Learning Scenarios of OTTA Algorithms}
\method{Stationary \vs Dynamic}
In contrast to vanilla OTTA \citep{wang2021tent} that assumes the test data comes from a stationary distribution, dynamic OTTA assumes a dynamically changing distribution including continual OTTA \citep{wang2022continual}, temporal OTTA \citep{gong2022note}, gradual OTTA \citep{dobler2023robust}, \tim{and practical OTTA \citep{yuan2023robust}.
A recent study \citep{marsden2024universal} delves into the realm of universal OTTA, a more complex setting where both domain non-stationarity and temporal correlation may coexist, with the specific test-time scenario often remaining unknown.}

\method{Data \vs Label shifts}
While the majority of OTTA methods concentrate on shifts in data distribution, some approaches \citep{yang2008non,royer2015classifier,wu2021online} investigate changes in label distribution.
Two interesting cases with online feedback are studied in \citep{royer2015classifier}, \ie, online feedback (the correct label is revealed to the system after prediction) and bandit feedback (the decision made by the system is correct or not is revealed).

Other differences between OTTA methods are the same as TTBA, \ie, \textbf{instance \vs batch}, \textbf{customized \vs on-the-fly}, and \textbf{single \vs multiple}.

\section{Applications \protect\footnote{
\tim{A table of commonly used datasets across various TTA applications is also provided in the GitHub repository.
}}}
\label{sec:appl}
\subsection{Image Classification}
The most common application of test-time adaptation is multi-class image classification.
Firstly, TTDA methods are commonly evaluated and compared on widely used DA datasets, including Digits, Office, Office-Home, VisDA-C, and DomainNet, as described in previous studies \citep{liang2020we,liang2021source,zhang2022divide}.
Secondly, TTBA and OTTA methods consider natural distribution shifts in object recognition datasets, \eg, corruptions in CIFAR-10-C, CIFAR-100-C, and ImageNet-C, natural renditions in ImageNet-R, misclassified real-world samples in ImageNet-A, and unknown distribution shifts in CIFAR-10.1, as detailed in previous studies \citep{sun2020test,schneider2020improving,wang2021tent,zhang2022memo}.
In addition, TTBA and OTTA methods are also evaluated in DG datasets such as VLCS, PACS, and Office-Home, as described in previous studies \citep{dinnocente2019learning,pandey2021generalization,iwasawa2021test,gan2023decorate}.

\subsection{Semantic Segmentation}
Semantic segmentation aims to categorize each pixel of the image into a set of semantic labels, which is a critical module in autonomous driving.
Many domain adaptive semantic segmentation datasets, such as GTA5-to-Cityscapes, SYNTHIA-to-Cityscapes, and Cityscapes-to-Cross-City, are commonly adopted to evaluate TTDA methods, as depicted in \citep{sivaprasad2021uncertainty,liu2021source,wang2022source}.
In addition to these datasets, BDD100k, Mapillary, and WildDash2, and IDD are also used to conduct comparisons for TTBA and OTTA methods, as shown in \citep{zou2022learning,bahmani2022adaptive}.
OTTA methods further utilize Cityscapes-to-ACDC and Cityscapes-to-Foggy\&Rainy Cityscapes for evaluation and comparison, as described in \citep{wang2022continual,volpi2022on}.

\subsection{Object Detection}
Object detection is a fundamental computer vision task that involves locating instances of objects in images.
While early TTA methods \citep{jamal2018deep,roychowdhury2019automatic} focus on binary tasks such as pedestrian and face detection, lots of current efforts are devoted to generic multi-class object detection.
Typically, many domain adaptive object detection tasks including Cityscapes-to-BDD100k, Cityscapes-to-Foggy Cityscapes, KITTI-to-Cityscapes, Sim10k-to-Cityscapes, Pascal-to-Clipart\&Watercolor are commonly used by TTDA methods for evaluation and comparison, as detailed in \citep{li2021free,huang2021model,li2022source_cvpr,sinha2023test}.
Additionally, datasets like VOC-to-Social Bikes and VOC-to-AMD are employed to evaluate TTBA methods \citep{dinnocente2020one,borlino2022self}.

\subsection{Beyond Vanilla Object Images}
\method{Medical images}
Medical image analysis is another important downstream field of TTA methods, \eg, medical image classification \citep{ma2022test,wang2022metateacher}, medical image segmentation \citep{he2021autoencoder,karani2021test}, and medical image detection \citep{liu2022source}.
Among them, medical segmentation attracts the most attention in this field.

\method{3D point clouds}
Nowadays, 3D sensors have become a crucial component of perception systems.
Many tasks for 2D images have been adapted for LiDAR point clouds, such as 3D object classification \citep{tian2022vdm}, 3D semantic segmentation \citep{saltori2022gipso}, and 3D object detection \citep{saltori2020sf}.

\method{Videos}
As mentioned above, TTBA and OTTA methods can address how to efficiently adapt an image model to real-time video data for problems such as depth prediction \citep{liu2023meta} and frame interpolation \citep{choi2021test}.
Besides, a few studies investigate the TTDA scheme for other video-based tasks including \tim{action recognition \citep{xu2022learning,huang2022relative,yi2023temporal,zeng2023exploring}, optical flow estimation \citep{ayyoubzadeh2023test} and object segmentation \citep{bertrand2023test}}.

\method{Multi-modal data}
Researchers also develop different TTA methods for various multi-modal data, \eg, RGB and audio \citep{plananamente2022test}, RGB and depth \citep{ahmed2022cross,shin2022mmtta}, RGB and motion \citep{huang2022relative}, \tim{and image-text pairs \citep{wen2024test}.
Furthermore, the development of multi-modal pre-trained models such as CLIP \citep{radford2021learning} enables image classification through image-to-text matching, gaining popularity among recent TTA methods \citep{samadh2023align,zhou2023test,ma2023swapprompt,zhao2024test}.}

\method{Face and body data}
Facial data is also an important application of TTA methods, such as face recognition \citep{zhang2022free}, face anti-spoofing \citep{wang2021self,liu2022source_eccv,zhou2022generative}, and expression recognition \citep{conti2022cluster}.
For body data, TTA methods also pay attention to tasks such as pose estimation \citep{zhang2020inference,kan2022self,ding2023maps} and mesh reconstruction \citep{guan2021bilevel,li2020online}.

\subsection{Beyond Vanilla Recognition Problems}
\method{Low-level vision}
TTA methods can be applied to low-level vision problems, \eg, image \tim{super-resolution \citep{park2020fast,deng2023efficient}}, image deblurring \citep{chi2021test}, and image dehazing \citep{liu2022towards}.
Besides, TTA is also introduced to image registration \citep{zhu2021test,hong2021deep}, inverse problems \citep{hussein2020image,darestani2022test}, and quality assessment \citep{liu2022source_arxiv}.

\method{Retrieval}
Besides classification problems, TTA can also be applied to kinds of retrieval scenarios, \eg, person re-identification \citep{wu2019distilled,xu2022mimic}, sketch-to-image retrieval \citep{sain2022sketch3t,paul2022ttt}, \tim{image-text matching \citep{zhou2023test}, and fair image retrieval \citep{kong2023mitigating}}.

\method{Generative modeling}
TTA method can also vary the pre-trained generative model for style transfer and data generation \citep{bau2019semantic,kim2022controllable,nitzan2022mystyle}.

\method{Defense}
Another interesting application is test-time adversarial defense \citep{shi2021online,yoon2021adversarial,alfarra2022combating}, which tries to generate robust predictions for possible perturbed samples.

\subsection{Natural Language Processing (NLP)}
The TTA paradigm is also studied in tasks of the NLP field, such as reading comprehension \citep{banerjee2021self}, question answering \citep{ye2022robust}, sentiment analysis \citep{zhang2021matching}, entity recognition \citep{wang2021efficient}, and aspect prediction \citep{ben2022pada}.
In particular, a competition~\footnote{\url{https://competitions.codalab.org/competitions/26152}} has been launched under data sharing restrictions, comprising two NLP semantic tasks \citep{laparra2021semeval}: negation detection and time expression recognition.

\subsection{Beyond CV and NLP}
\method{Graph data}
For graph data (\eg, social networks), TTA methods are evaluated and compared on either graph classification \citep{wang2022testtime} or node classification \citep{jin2023empowering}.

\method{Speech processing}
As far, there have been three TTA methods, \ie, audio classification \citep{boudiaf2023in}, speaker verification \citep{kim2022variational} and speech recognition \citep{lin2022listen}.

\method{Miscellaneous signals}
TTA methods have been also validated on other types of signals, \eg, radar signals \citep{cao2021towards}, EEG signals \citep{lee2023source}, and vibration signals \citep{jiao2022source}.

\method{Reinforcement learning}
Some TTA methods \citep{hansen2021self,liu2023learning} also address the generalization of reinforcement learning policies across different environments.

\subsection{Evaluation}
As the name suggests, TTA methods should evaluate the performance of test data after test-time optimization immediately.
However, there are different protocols for evaluating TTA methods in the field, making a rigorous evaluation protocol important.
Firstly, some TTDA works, particularly for domain adaptive semantic segmentation \citep{sivaprasad2021uncertainty,wang2022source} and classification on DomainNet, adapt the source model to an unlabeled target set and evaluate the performance on the test set that shares the same distribution as the target set. 
However, this in principle violates the setting of TTA, although the performance on the test set is always consistent with that of the target set. 
\emph{We suggest that such SFDA methods report the performance on the target set at the same time.}
Secondly, some TTDA works such as BAIT \citep{yang2021casting} offer an online variant, but such online TTDA methods differ from OTTA in that the evaluation is conducted after one full epoch.
\emph{We suggest online TTDA methods change the name to ``one-epoch TTDA" to avoid confusion with OTTA methods.}
Thirdly, for continual TTA methods \citep{wang2022continual,niu2022efficient}, the evaluation of each mini-batch is conducted before optimization on that mini-batch.
This manner differs from the standard evaluation protocol of OTTA \citep{sun2020test} where optimization is conducted ahead of evaluation.
\emph{We suggest that continual TTA methods follow the same protocol as vanilla OTTA methods.}
 
\section{Emerging Trends and Open Problems}
\label{sec:future}
\subsection{Emerging Trends}
\method{Diverse downstream fields}
Even most existing efforts in the TTA field have been devoted to visual tasks such as image classification and semantic segmentation, a growing number of TTA methods are now focusing on other understanding problems over video data \citep{xu2022learning}, multi-modal data \citep{shin2022mmtta}, and 3D point clouds \citep{saltori2022gipso}, as well as regression problems like pose estimation \citep{ding2023maps}.


\method{Open-world adaptation}
Existing TTA methods always follow the closed-set assumption; however, a growing number of TTDA methods \citep{liang2021umad,yang2023one,qu2023upcycling} are beginning to explore model adaptation under an open-set setting.
A recent OTTA method \citep{yang2023auto} further focuses on the performance of out-of-distribution detection tasks at test time.
Besides, for large distribution shifts, it is challenging to perform effective knowledge transfer by relying solely on unlabeled target data, thus several recent works \citep{li2022source_mm,kothandaraman2023salad} also introduce active learning to involve humans in the loop.

\method{Memory-efficient continual adaptation}
In real-world applications, test samples may come from a continually changing environment \citep{wang2022continual,niu2022efficient}, leading to catastrophic forgetting.
To reduce memory consumption while maintaining accuracy, recent works \citep{song2023ecotta,hong2023mecta} propose different memory-friendly OTTA solutions for resource-limited end devices.

\method{On-the-fly adaptation}
The majority of existing TTA methods require a customized pre-trained model from the source domain, bringing the inconvenience for instant adaptation.
Thus, fully test-time adaptation \citep{wang2021tent}, which allows adaptation with an on-the-fly model, has attracted increasing attention.

\method{Foundation models}
\tim{Large language models like GPT have attracted widespread attention due to their surprisingly strong ability in various tasks. 
Given a query to a language model, a recent work \citep{hardt2024test} performs test-time training by fine-tuning the model based on its retrieved nearest neighbors.
Over the past two years, there has been a growing number of TTBA methods \citep{shu2022test,feng2023diverse,samadh2023align,zhou2023test,zhao2024test,yoon2024c} developed that leverage vision-language models, such as CLIP \citep{radford2021learning}, to enhance the zero-shot generalization.
Meanwhile, some studies have focused on CLIP adaptation under the OTTA scenario \citep{ma2023swapprompt,karmanov2024efficient} as well as the TTDA setting \citep{tanwisuth2023pouf,hu2024reclip}.
Additionally, several recent studies \citep{feng2023diverse,prabhudesai2023test} have explored leveraging large-scale generative models, such as Stable Diffusion \citep{rombach2022high}, for developing TTA methods.}

\subsection{Open Problems}
\method{Theoretical analysis}
While most existing works focus on developing effective TTA methods to obtain better empirical performance, the theoretical analysis \tim{of when and why TTA works remains an open problem.
Several TTA methods have provided theoretical results on specific designs under linear models such as gradient descent with pseudo-labels \citep{wang2023towards} and auxiliary self-supervision \citep{sun2020test}.
One recent work \citep{gui2024active} conducts an in-depth theoretical analysis based on learning theories and mainly explores how can significant distribution shifts be effectively addressed under the online TTA setting.
We believe that more rigorous analyses, especially on deep learning models, can provide deeper insights} and inspire the development of new TTA methods.

\method{Benchmark and validation}
\tim{Recently, several new benchmarks \citep{yu2023benchmarking,press2023rdumb,wang2023search} are proposed to fairly evaluate various TTA methods.
For example, the vision transformer (ViT) architecture is further employed for online TTA methods in \citep{wang2023search}, and a new dataset is developed to testify online TTA methods under continuously changing corruptions \citep{press2023rdumb}.
}
However, as there does not exist a labeled validation set, validation also remains a significant and unsolved issue for TTA methods. 
\tim{As noted in \citep{zhao2023pitfalls}, evaluations of TTA methods have often been conducted unfairly. 
Existing studies frequently determine hyper-parameters through grid search on the test data, which is not feasible in real-world applications.
To address this issue, a recent benchmark \citep{yu2023benchmarking} has proposed a fixed validation strategy with a predetermined online batch order.
It selects the optimal hyper-parameters based on the first one of the adaptation tasks for all the tasks.}
In the future, a benchmark can be built where a labeled validation set and an unlabeled test set exist at test time, providing a more realistic evaluation scenario for TTA methods.

\method{New applications}
Tabular data \citep{borisov2022deep} in vectors of heterogeneous features is essential for industrial applications, and time series data \citep{ragab2023adatime} is predominant in real-world applications like healthcare and manufacturing.
So far, limited prior work has explored TTA in the context of tabular or time series data, despite their importance and prevalence in real-world scenarios.
\tim{When it comes to adapting to tabular data, deep learning models have generally underperformed compared to tree-based models such as XGBoost and random forests \citep{shwartz2022tabular,grinsztajn2022tree}. 
Therefore, it would be interesting to investigate how TTA methods developed primarily for deep learning models can be applied and perform when used with tree-based models for tabular data scenarios.}

\method{Trustworthiness}
Current TTA methods focus more on robustness under distribution shifts while ignoring other goals of trustworthy machine learning \citep{eshete2021making}, \eg, fairness, security, privacy, and explainability.
\tim{Regarding class-wise fairness, the adapted model's performance may vary considerably across different categories in the target domain. 
However, existing TTA methods have not thoroughly investigated the worst-class accuracy for classification tasks.
As for security, in the TTDA setting, the source provider could potentially be a malicious actor who inserts backdoors into the pre-trained model \citep{sheng2023adaptguard}. 
This could enable the attacker to then target the model adapted by the end user using the same embedded backdoor triggers.}
\tim{Furthermore, another important issue with existing TTA methods is their tendency towards overconfidence, which undermines the reliability of their predictions \citep{kim2023reliable,yoon2024c}.}
 
\section{Conclusion}
Learning to adapt a pre-trained model to unlabeled data under distribution shifts is an emerging and critical problem in the field of machine learning. 
This survey provides a comprehensive review of three related topics: test-time domain adaptation, test-time batch adaptation, and online test-time adaptation.
These topics are unified as a broad learning paradigm of test-time adaptation (TTA).
For each topic, we first introduce its definition and a new taxonomy of advanced algorithms.
Additionally, we provide a review of applications related to test-time adaptation, as well as an outlook of emerging research trends and open problems. 
We believe that this survey will assist both newcomers and experienced researchers in better understanding the current state of research in TTA under distribution shifts.

\section*{Acknowledgements}
The authors sincerely thank the editor and
anonymous reviewers for their constructive comments on this work.

\small
{
\bibliographystyle{unsrt}
\bibliography{ref,p1,p2,p3}
}

\end{document}